\documentclass{article}

\usepackage{microtype}
\usepackage{graphicx}
\usepackage{subfigure}
\usepackage{booktabs} 

\usepackage{hyperref}
\usepackage{float}

\usepackage[accepted]{icml2020}

\newcommand{\x}{\mathbf{x}}
\newcommand{\fx}{f(\mathbf{x})}
\newcommand{\fmax}{f_{\text{max}}}

\usepackage{todonotes}
\usepackage{amsmath,amsthm,mathtools,amssymb}
\usepackage{microtype}
\usepackage{xspace}
\usepackage{todonotes}
\usepackage{enumitem}
\usepackage{multirow}

\newcommand{\eg}{\textit{e.g.}\xspace}

\newcommand{\expect}{\mathbb E}
\newcommand{\propose}{\textup{propose}\xspace}
\newcommand{\fit}{\textup{fit}\xspace}

\newcommand{\figref}[1]{Figure~\ref{#1}}

\newcommand{\secref}[1]{Section~\ref{#1}}

\definecolor{linkcolor}{rgb}{0,0.08,0.45}
\newcommand{\sfigref}[2]{Suppl. Figure~\textcolor{linkcolor}{#2}}

\newcommand{\stabref}[2]{Suppl. Table~\textcolor{linkcolor}{#2}}
\newcommand{\ssecref}[2]{Suppl. Section~\textcolor{linkcolor}{#2}}

\newcommand{\pbo}{P3BO\xspace}  
\newcommand{\apbo}{Adaptive-P3BO\xspace}  
  
\definecolor{diagram-red}{rgb}{0.84, 0.19, 0.153}
\definecolor{diagram-blue}{rgb}{0.27, 0.459, 0.706}
\definecolor{diagram-yello}{rgb}{0.992, 0.682, 0.38}
\icmltitlerunning{Population-Based Black-Box Optimization for Biological Sequence Design}

\definecolor{darkblue}{rgb}{0.0,0.0,0.55}
\hypersetup{
  colorlinks = true,
  citecolor  = darkblue,
  linkcolor  = darkblue,
  citecolor  = darkblue,
  filecolor  = darkblue,
  urlcolor   = darkblue,
}
\usepackage[font=small,labelfont=bf]{caption}
\usepackage{bm}

\usepackage[noend]{algpseudocode}

\usepackage{etoolbox}
\usepackage{tikz}
\usetikzlibrary{tikzmark}
\usetikzlibrary{calc}

\errorcontextlines\maxdimen

\newcommand{\ALGtikzmarkcolor}{black}
\newcommand{\ALGtikzmarkextraindent}{4pt}
\newcommand{\ALGtikzmarkverticaloffsetstart}{-1ex}
\newcommand{\ALGtikzmarkverticaloffsetend}{-.5ex}
\makeatletter
\newcounter{ALG@tikzmark@tempcnta}

\newcommand\ALG@tikzmark@start{%
    \global\let\ALG@tikzmark@last\ALG@tikzmark@starttext%
    \expandafter\edef\csname ALG@tikzmark@\theALG@nested\endcsname{\theALG@tikzmark@tempcnta}%
    \tikzmark{ALG@tikzmark@start@\csname ALG@tikzmark@\theALG@nested\endcsname}%
    \addtocounter{ALG@tikzmark@tempcnta}{1}%
}

\def\ALG@tikzmark@starttext{start}
\newcommand\ALG@tikzmark@end{%
    \ifx\ALG@tikzmark@last\ALG@tikzmark@starttext
    \else
        \tikzmark{ALG@tikzmark@end@\csname ALG@tikzmark@\theALG@nested\endcsname}%
        \tikz[overlay,remember picture] \draw[\ALGtikzmarkcolor] let \p{S}=($(pic cs:ALG@tikzmark@start@\csname ALG@tikzmark@\theALG@nested\endcsname)+(\ALGtikzmarkextraindent,\ALGtikzmarkverticaloffsetstart)$), \p{E}=($(pic cs:ALG@tikzmark@end@\csname ALG@tikzmark@\theALG@nested\endcsname)+(\ALGtikzmarkextraindent,\ALGtikzmarkverticaloffsetend)$) in (\x{S},\y{S})--(\x{S},\y{E});%
    \fi
    \gdef\ALG@tikzmark@last{end}%
}

\apptocmd{\ALG@beginblock}{\ALG@tikzmark@start}{}{\errmessage{failed to patch}}
\pretocmd{\ALG@endblock}{\ALG@tikzmark@end}{}{\errmessage{failed to patch}}
\makeatother

\begin{document}

\twocolumn[
\icmltitle{Population-Based Black-Box Optimization for Biological Sequence Design}

\icmlsetsymbol{equal}{*}

\begin{icmlauthorlist}
\icmlauthor{Christof Angermueller}{goo}
\icmlauthor{David Belanger}{goo}
\icmlauthor{Andreea Gane}{goo}
\icmlauthor{Zelda Mariet}{goo}
\icmlauthor{David Dohan}{goo}
\icmlauthor{Kevin Murphy}{goo}
\icmlauthor{Lucy Colwell}{goo,uoc}
\icmlauthor{D Sculley}{goo}
\end{icmlauthorlist}

\icmlaffiliation{goo}{Google Research}
\icmlaffiliation{uoc}{University of Cambridge}

\icmlcorrespondingauthor{Christof Angermueller}{christofa@google.com}

\icmlkeywords{Machine Learning, Black-box optimization, Protein design, Protein engineering}

\vskip 0.3in
]



\printAffiliationsAndNotice{}  

\begin{abstract}
The use of black-box optimization for the design of new biological sequences is an emerging research area with potentially revolutionary impact. The cost and latency of wet-lab experiments requires methods that find good sequences in few experimental rounds of large batches of sequences --- a setting that off-the-shelf black-box optimization methods are ill-equipped to handle. We find that the performance of existing methods varies drastically across optimization tasks, posing a significant obstacle to real-world applications. To improve robustness, we propose Population-Based Black-Box Optimization (\pbo), which generates batches of sequences by sampling from an ensemble of methods. The number of sequences sampled from any method is proportional to the quality of sequences it previously proposed, allowing \pbo to combine the strengths of individual methods while hedging against their innate brittleness. Adapting the hyper-parameters of each of the methods online using evolutionary optimization further improves performance. Through extensive experiments on \emph{in-silico} optimization tasks, we show that \pbo outperforms any single method in its population, proposing  higher quality sequences as well as more diverse batches. As such, \pbo and Adaptive-\pbo are a crucial step towards deploying ML to real-world sequence design.
\vspace{-1em}
\end{abstract}

\section{Introduction}
\label{sec:intro}
The ability to design new protein or DNA sequences with desired properties would revolutionize drug discovery, healthcare, and agriculture. However, this is a particularly challenging optimization problem: the space of sequences is discrete and exponentially large; evaluating the fitness of proposed sequences requires costly wet-lab experiments; and not just one but a {\em diverse} set of high-quality sequences must be discovered to improve the chance of a candidate surviving downstream screening (\eg, for toxicity).  

The problem is not insurmountable, however, due to modern experimental technology, where hundreds to thousands of sequences can be evaluated in parallel.  This forms the basis for directed evolution, a form of human-guided local evolutionary search ~\citep{arnold1998design}, and several ML-based methods (\secref{sec:rw-ml}). Additionally, development of ML methods suitable to guide sequence design can be aided by working {\em in-silico}, instead of relying on wet-lab processes during algorithmic development.

\begin{figure*}[!h]
\begin{center}
\vspace{-0.5em}
\includegraphics[width=.85\textwidth]{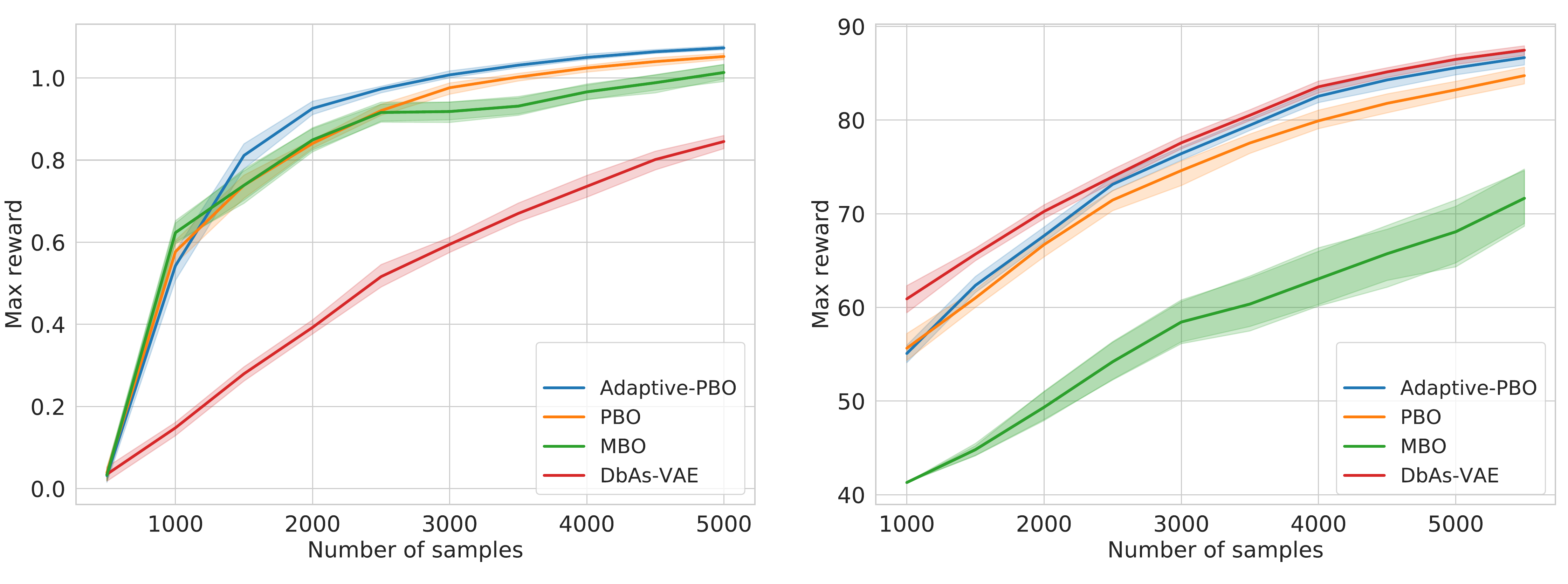}
\vspace{-1.2em}
\caption{Motivating example: comparison of optimization trajectories for baseline ML methods and \pbo on optimization problems (Section~\ref{sec:exp-problems}) with qualitatively-different  functional forms: PdbIsing (left) and PfamHMM (right). The rank ordering of performance among the baseline methods for these two seemingly similar problems is inverted, showing the potential brittleness of existing approaches for biological sequence design.}
\label{fig:motivation}
\end{center}
\vspace{-1.6em}
\end{figure*}

In this paper, we introduce several \textit{in-silico} design problems upon which we evaluate such ML methods. Unfortunately, we find that popular optimization methods are particularly sensitive to hyper-parameter choice and can have strong inductive biases that allow them to excel at some problems but perform poorly on others (Figure~\ref{fig:motivation}).  This lack of robustness is a serious concern for practical application of these methods, which could cause wet-lab experiments to fail.  We further find that several existing methods are ill-suited to generating diverse batches of sequences. Instead of using the  batch size efficiently, methods tend to generate very similar sequences.

To improve robustness and sequence diversity, we introduce Population-Based Black-box Optimization (\pbo). \pbo draws inspiration from portfolio algorithms~\citep{leyton-brown_portfolio_2003, tang_population-based_2014} for numerical optimization: instead of generating a batch of sequences using a single potentially brittle algorithm, \pbo samples sequences from a \emph{portfolio} of algorithms, allocating budget to each algorithm based on the quality of its past proposed sequences. 

To our knowledge, \pbo is the first approach to leverage an ensemble of optimizers for \emph{batched} optimization, a crucial characteristic of wet-lab optimization loops. We show that batching offers a dimension along which ensembling yields significant gains.

Notably, samples acquired by one algorithm are shared with all other algorithms, allowing each of them to learn from data acquired by all. By combining the strengths of multiple algorithms, \pbo hedges against the risk of choosing an unsuitable optimization algorithm for the problem at hand~(\secref{sec:robustness}). Sampling sequences from multiple algorithms additionally allows \pbo to produce more diverse batches and find distinct optima faster. We employ a heterogenous population, consisting of global model-based optimizers based on discriminative and generative models along evolutionary strategies. Finally, we further improve \pbo by introducing a variant, Adaptive-\pbo, which adapts the hyper-parameters of the algorithms themselves on the fly using evolutionary search.

We evaluate \pbo and Adaptive-\pbo empirically on over 100 batched black-box optimization problems, and show that \pbo and Adaptive-\pbo are considerably more robust, generate more diverse batches of sequences, and find distinct optima faster than any single method in their population. Adaptive-\pbo improves upon \pbo results, and furthermore is able to recover from a poor initial population of methods. Our contributions are as follows:
\vspace{-.5em}
\begin{itemize}[leftmargin=*]
    \item We introduce new in-silico optimization problems for benchmarking biological sequence design methods.
    \item We evaluate state-of-the-art sequence design methods across these problems, bringing to light two significant shortcomings of existing methods: (a) lack of generalization across similar problem classes, and (b) sub-optimal use of the large batch sizes crucial to wet-lab settings.
    \item We introduce \pbo: a population-based optimization framework for discrete batched black-box function optimization that ensembles over  algorithms to hedge against brittleness and improve diverse sequence discovery. 
    \item We introduce Adaptive-\pbo, an extension of \pbo that tunes the hyper-parameters of population members using evolutionary search, yielding further improvements upon \pbo.
\end{itemize}
\section{Problem Setting and Motivation}
\label{sec:problem-setting}

\begin{figure*}
    \centering
    \def\svgwidth{\textwidth}
    \vspace{-1.1em}
    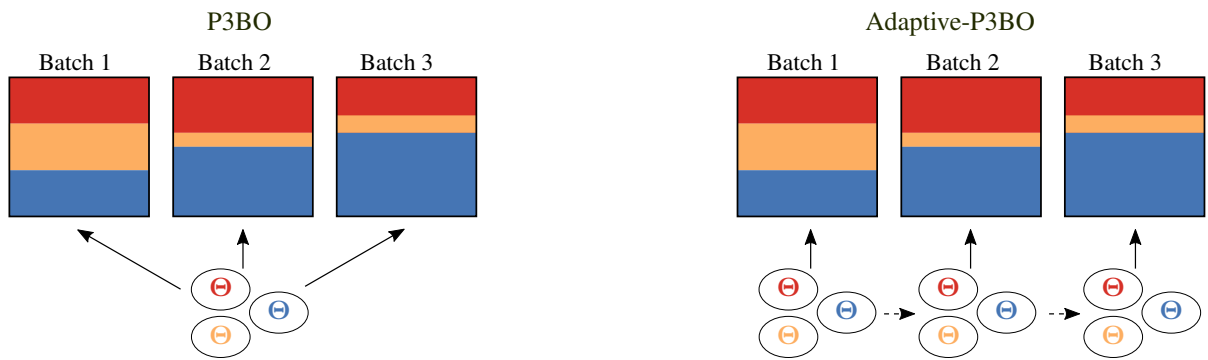
    \vspace{-1.7em}
    \caption{\small \pbo (left) adjusts the fraction of each batch that is allocated to a particular algorithm in its population based on the quality of sequences that the algorithm proposed in the past. Adaptive-\pbo (right) additionally adapts the hyper-parameters $\Theta$ of algorithms by evolutionary search.}
    \label{fig:diagram}
    \vspace{-1.2em}
\end{figure*}

We define sequences $\x$ as elements of $\mathcal{V}^L$, where $\mathcal{V}$ is a finite vocabulary (for DNA, $|\mathcal{V}| = 4$; for proteins, $|\mathcal{V}| = 20$) and $L$ is the sequence length. For variable length sequences, we assume that sequences are padded to length $L$ by an end-of-sequence token. 

Sequence design aims to maximize a function $f: \mathcal{V}^L \to \mathbb R$, which can be evaluated on batches of sequences $\mathcal{X} \subseteq \mathcal{V}^L$ size $B=|\mathcal{X}|$, but only a limited number of times $T$. 

\subsection{Algorithm Requirements}
Most discrete black-box optimization methods have associated hyper-parameters. Throughout the paper, an \emph{algorithm} $A$ refers to an instance of a particular method class (\eg, evolutionary search), which is instantiated by a specific hyper-parameter configuration (\eg, mutation rate for evolutionary search). As \pbo ensembles heterogeneous algorithms, including global model-based optimizers and local search strategies, we make the following assumptions about the interface of algorithms.

$A.\fit(\mathcal{X}, \mathcal{Y})$ updates its internal state (\eg, ML model) using a batch of sequences $\mathcal{X}$ and their objective values $\mathcal{Y} = \{\fx \mid \x \in \mathcal{X}\}$. Next, $A.\propose()$ suggests a single sequence to be evaluated next. We require that $\fit(\mathcal{X}, \mathcal{Y})$ can leverage data obtained by other algorithms, which prohibits the use of on-policy RL methods.

\subsection{Robustness of Black-Box Optimizers}
\label{sec:robustness}
\figref{fig:motivation} demonstrates the lack of robustness of existing optimization methods (\secref{sec:exp-methods}) on two representative in-silico optimization problems (\secref{sec:exp-problems}). On PDBIsing (left), model-based optimization (MBO), an optimizer based on a discriminative model of $\fx$, performs well, and DbAs-VAE, which employs a generative model, struggles. On PfamHMM, their relative performance reverses order. This is in large part due to differences in the compatibility of methods' inductive biases with the form of $\fx$. On PDBIsing, the objective function is the sum of local terms at each sequence position and long-range pairwise interactions between positions. Given limited samples, the discriminative model can estimate both of these terms, and propose high-quality sequences accordingly. On PfamHMM, the objective function is the likelihood of a generative model fit to data with insertions and deletions. A optimization method based on a generative model can capture such variability. In real applications, it is crucial that methods are robust to the structure of $\fx$.  \pbo, our proposed population-based approach, performs at least as well as the best algorithm in its population and often exceeds its performance. 

\section{Population-Based Optimization for Batched Sequence Design}

We introduce \pbo, a robust black-box optimization method that constructs batches of sequences with which to query $\fx$ using a population of heterogeneous optimization algorithms. By sharing data between algorithms, algorithms benefit from each other's distinct exploration strategies.

\subsection{Method Summary}
The high-level structure of \pbo is summarized in Algorithm~\ref{alg:pbo} and \figref{fig:diagram}. The input is an initial population $\mathcal{P}^{0} = \{A_1,..., A_N\}$ of $N$ constituent algorithms, which can be sampled from a distribution over algorithms or chosen using prior knowledge. At experimental round $t$, \pbo constructs a batch $\mathcal{X}^{t}$ of $B$ sequences by iteratively sampling algorithm an $A_i$ from a categorical distribution parameterized by $p^{t}$. The sampled algorithm $A_i$ is then used to propose a sequences $\x$, which is added to the batch $\mathcal{X}^t$. Note that the constructed batch is  a set $\mathcal{X}^t$ of \emph{unique} sequences, assuring that $\fx$ is not evaluated on identical sequences, which would be a waste of resources.

\pbo weights algorithms proportional to the quality of sequences that they found in the past, sampling more sequences from algorithms that found higher quality sequences. This is done by computing a reward $r^{t}_i$ for each algorithm $A_i$ at each step $t$ and adjusting the probability of sampling from $A_i$ based on $r^{t}_i$. To evaluate the performance of each algorithm in the population, \pbo keeps track of which algorithms proposed each sequence in $\mathcal{X}^{t}$  using subsets $\mathcal{X}^{t}_i$. If two algorithms $A_i$ and $A_j$ propose the same sequence $\x \in \mathcal{X}^t$,  $\x$ is added to both $\mathcal{X}^t_i$ and $\mathcal{X}^t_j$.

After generating batch $\mathcal{X}^{t}$, the target function $\fx$ is evaluated, yielding observations $\mathcal{Y}^t$. These are used to compute \textit{rewards} for each method (\secref{sec:credit-scores}), which are used in turn to compute sampling probabilities $p^{t+1}$ for the next round. Finally, the $\text{fit}()$ method of all algorithms is called on the extended data $(\mathcal{X}, \mathcal{Y})$ to update their internal state. The fitting step itself is algorithm-specific, and can entail time-consuming steps such as fitting discriminative or generative models. 

\begin{algorithm}[h]
  \caption{\pbo}
  \label{alg:pbo}
\begin{algorithmic}
  \State {\bfseries Input:} Population $\mathcal{P} = \{A_1, \ldots, A_N\}$
  \State {\bfseries Input:} Softmax temperature $\tau > 0$.
  \State {\bfseries Input:} Initial sampling weights $p^1$
  \State $\mathcal{X}, \mathcal{Y} = \emptyset, \emptyset$ \Comment{Sequences and labels} \vphantom{$\int^X$} 
  
  \For{$t=1$ {\bfseries to} $T$}
  \State $\mathcal{X}^{t} = \emptyset$
  \State $\mathcal{X}^{t}_1, \ldots, \mathcal{X}^{t}_N = \emptyset, \ldots, \emptyset$
  \While{$|\mathcal{X}^{t}| \le B$}
    \State $i \sim \text{Categorical}(p^{t})$
    \State $x = A_i.\text{propose}()$
    \State $\mathcal{X}^{t} \leftarrow \mathcal{X}^{t} \cup \{x\}$ \Comment{Only add $x$ if novel}
    \State $\mathcal{X}^{t}_i \leftarrow \mathcal{X}^{t}_i \cup \{x\}$ \Comment{Only add $x$ if novel}
  \EndWhile
  \State $\mathcal{Y}^{t} = \{f(x) \mid x \in \mathcal{X}^{t}\}$
  \State $\mathcal{X}, \mathcal{Y} \leftarrow \mathcal{X} \cup \mathcal{X}^{t}, \mathcal{Y} \cup \mathcal{Y}^{t}$
  \State $r^{t} = \text{get\_rewards}\big(\mathcal{X}, \mathcal{Y}, \{\mathcal{X}^{t}_i\}_i\big)$ \Comment{Eq.~\ref{eq:rewards}}
  \State $s^t = \text{decayed\_rewards}(r^t)$ \Comment{Eq.~\ref{eq:decayed-rewards}}
  \If{Adaptive \pbo}
    \State $\mathcal{P}, s^{t} = \text{adapt}(\mathcal{P}^{t}, s^{t})$ \Comment{Alg.~\ref{alg:adapt}}
  \EndIf
  \State $p^{t+1} = \text{softmax}(\hat s^t / \tau)$ \Comment{Eq.~\ref{eq:sampling-prob}}
  \For{$A_i \in \mathcal{P}$}
  \State $A_i.\text{fit}(\mathcal{X}, \mathcal{Y})$
  \EndFor
  \EndFor
  \State {\bfseries return} $X$
\end{algorithmic}
\end{algorithm}
 
\subsection{Selecting from a Population of Algorithms}
\label{sec:credit-scores}
To generate high-quality sequences consistently across optimization rounds and different optimization problems, \pbo must adjust each constituent algorithm's contribution over time. At round $t$, \pbo observes all objective function values for sequences $\mathcal{X}^{t}_i$ that were proposed by algorithm $A_i$. These observations are converted into a per-algorithm reward $r^t_i$. Doing so is challenging because the the qualities of the sequences from each algorithm are correlated, since the algorithms share observations between optimization rounds. 

In our experiments, we used the improvement of $\fx$ relative to $\fmax = \max \{f(x) \mid x \in \mathcal{X}\}$, the maximum of $\fx$ of all previous rounds:
\begin{equation}
r^{t}_i = \frac{\max \{\fx \mid \x \in \mathcal{X}^t_i\} - \fmax}{\fmax}.
\label{eq:rewards}
\end{equation}
Future work should consider alternative reward functions to~\eqref{eq:rewards}, such as novelty search~\cite{lehman2008exploiting}.

As \pbo aims to address lack of robustness in existing methods, we prefer algorithms that consistently propose good sequences. To do so, the probability $p_i$ of sampling algorithm $A_i$ depends not only on its reward at time $t$ but the sum of exponentially decayed rewards via a \emph{credit score} $s^t_i$:
\begin{align}
  s^{t}_i &= \sum_{t’ \le t} r^{t’}_i \gamma^{t - t’}, \label{eq:decayed-rewards}\\
  p_i &= \frac{\exp(\hat s_i / \tau)}{\sum\nolimits_j \exp(\hat s_j/\tau)},  \label{eq:sampling-prob}
\end{align}
The decay rate $\gamma$ trades-off past and present improvements, assigning higher credit scores to algorithms that improve $\fx$ consistently across optimization rounds. $\hat s_i$ are min-max normalized values of the credit scores, which ensures that the $p_i$ are independent of the scale of the rewards. The hyper-parameter $\tau$ controls the entropy of the distribution, effectively trading off the exploration and exploitation of algorithms. If $\tau$ is set to a high value, sequences are sampled uniformly from the population, regardless of their rewards. 

\subsection{Adaptive Population-Based Optimization}
\label{sec:adaptive-pbo}
Although the credit assignment and selection strategy described in Section~\ref{sec:credit-scores} increases robustness by sampling few or no sequences from poorly performing methods in the population, it is limited to the set of algorithms that are already in the population. For example, the hyper-parameters of algorithms in the population can be sub-optimal for a particular problem, which upper-bounds the performance of \pbo. We address this limitation by optimizing hyper-parameters of algorithms in the population online by evolutionary search (Algorithm~\ref{alg:adapt}).

We first select the set $\mathcal{S}$ of algorithms with the top-$q$ credit scores from $\mathcal{P}$, where $q$ is a quantile cut-off. We then use tournament selection \cite{miller1995genetic} to select $k=2$ parent algorithms from the pool of survivors $\mathcal{S}$, and recombine their hyper-parameters. If the parents belong to different classes of algorithms and their hyper-parameters are incompatible, we select one of them randomly. Otherwise, we crossover their hyper-parameters with some crossover rate. Finally, we mutate the resulting hyper-parameters with some mutation rate by either resampling hyper-parameters values from a prior distribution, or scaling them by a constant. 

\begin{algorithm}[h]
  \caption{Adaptation of population members}
  \label{alg:adapt}
\begin{algorithmic}
  \State {\bfseries Input:} Population of algorithms $\mathcal{P} = \{A_1, \ldots, A_N\}$
  \State {\bfseries Input:} Algorithm scores $s = \{s_1, \ldots, s_N\}$
  \State {\bfseries Input:} Quantile cutoff $q$
  \State $S = \{A_i \in \mathcal{P} \mid f_i \ge q\}$
  \State $\widetilde{\mathcal{P}}, \widetilde s = \emptyset, \emptyset$
  \For{$i=1$ {\bfseries to} $N$}
    \State parents $=$ tournament\_select$(\mathcal{S})$
    \State $A_i, s_i = \text{recombine(parents)}$
    \State $\widetilde A_i =$ mutate$(A_i)$
    \State $\widetilde{\mathcal{P}} \leftarrow \widetilde{\mathcal{P}} \cup \{\widetilde A_i\}$
    \State $\widetilde s \leftarrow \widetilde s \cup \{s_i\}$
  \EndFor
  \State {\bfseries return} $\widetilde{\mathcal{P}}, \widetilde s$
\end{algorithmic}
\end{algorithm}

\vspace{-1em}
\section{Related Work}
\label{sec:rw-ml}

Our work draws from related research in both ML-guided sequence design and population-based optimization. 
\vspace{-1em}
\paragraph{ML for Sequence Design.} Methods based on generative models seek to maximize the expected value $\expect_{p(\x)}[f(\x)]$ of the objective function $f(\x)$ when sampling sequences $\x$ from a distribution $p(x)$ that is parameterized, for example, using a RNN. Most approaches for maximizing this expectation can be seen as instances of the cross-entropy method~\cite{de2005tutorial, neil2018exploring,brookes2018design,gupta2018feedback,brookes2019view}. At each round, the distribution is trained to maximize the likelihood of high-reward sequences seen so far. The next batch is sampled from this distribution.

Throughout the paper, we use `model-based optimization' to refer to machine learning approaches that employ a discriminative surrogate model $\hat{f}(\x)$ that approximates $\fx$. The surrogate is converted into an \textit{acquisition function}, which is optimized to propose the next batch of sequences~\cite{hashimoto2018derivative,wang2019synthetic,yang2019machine,de2019designing,liu2019antibody,sample2019human,wu2019machine, biswas2020low}. 

Approximating $f(\x)$ by a surrogate $\hat{f}(\x)$ has two advantages. First, $\hat{f}(\x)$ is inexpensive to evaluate unlike $f(x)$, which can require costly wet-lab experiments. Second, knowledge of the structure of $\hat{f}(\x)$ or its gradients can be used to guide the optimization of the acquisition function. Existing methods differ in the form of $\hat{f}(\x)$, the type of the acquisition function (\eg, expected improvement~\cite{mockus1978application}), and the optimization of the acquisition.

Recently, \citet{Angermueller2020RLDesign} proposed a hybrid approach for sequence design, where a generative policy is updated using model-based RL if the surrogate discriminative model is accurate, and model-free RL otherwise. 

\vspace{-.5em}
\paragraph{Population-Based Optimization.} Ensembling is a common strategy to produce robust algorithms by combining diverse algorithms that have individual weaknesses. Ensembles of evolutionary and swarm algorithms have been proposed for non-batched optimization, including multi-strategy methods~\cite{du2008multi}, portfolio algorithms~\citep{leyton-brown_portfolio_2003, tang_population-based_2014}, and hyper-heuristics~\citep{burke2013hyper}. Existing approaches surveyed in~\citet{wu2019ensemble} can be categorized into low-level ensembles, which agglomerate different instances of the same class of algorithm such as different mutation operators for evolutionary search, and high-level ensembles, which operate over algorithms belonging to heterogeneous families. \pbo belongs to the later category.

Ensemble optimizers differ in how constituent algorithms are selected over time, also known as adaptive operator selection (AOS)~\cite{maturana2009extreme,fialho2010toward,li2013adaptive}. AOS first defines the credit score of operators based on their past rewards and then selects operators based on their score. The average reward, relative reward improvement, or sum of ranks are common credit assignment strategies. \citet{wu2019ensemble} includes a review of such operator selection techniques, which tend to resemble the popular weighted majority method~\cite{littlestone1989weighted} for multi-armed bandits. AOS is not a bandit problem, however, since the action at time $t$ impacts the rewards that different algorithms experience at later steps.

Evolutionary reinforcement learning~\cite{pourchot2018cem,khadka2018evolution} adapts a population of agents over time and exchanges observations between them. However, it is a low-level ensemble, combining homogeneous RL agents, and it performs non-batched, continuous optimization in the space of agents' parameters. Population-based training (PBT)~\cite{jaderberg2017population} jointly optimizes the weights and hyper-parameters of neural networks. However, the optimization is on a static training set, instead of data collected on-the-fly. AlphaStar applies population-based training to evolve a set of agents in a multi-agent RL setting \cite{vinyals2019grandmaster}.

\section{In-Silico Benchmarking Problems}
\label{sec:exp-problems}

Before deploying optimization methods on expensive wet-lab experiments, it is crucial to analyze them using in-silico surrogates. This section describes the benchmark problems that we have used to study the strengths and weaknesses of different methods. See \ssecref{sec:app-problems}{A} for further details.

Creating realistic problems is challenging because protein fitness landscapes have not been experimentally well-characterized, and understanding their properties is an open research question. However, we can create problems with a range of functional forms containing elements that appear in real landscapes. Our problems fall into four categories: (1) exhaustive wet-lab measurements of all sequences of a small search space, (2) regressors fit to wet-lab measurements for a subset of sequences from  problems with larger search spaces, (3) neural networks with random weights, and (4) statistical models for protein sequence evolution fit using experimental data. Problems vary in the size of their search space, the number of initial samples provided to optimizers, and the sensitivity $\fx$ to shifts, insertions, and deletions of $\x$. See Section~\ref{sec:exp-problems} for more details about individual optimization problems.

\textbf{TfBind8:}
\citet{barrera2016survey} measured the binding activity between a variety of human transcription factors and every possible length-8 DNA sequence. For each transcription factor, the optimization goal is to identify DNA sequences that maximize the binding activity score.

\textbf{TfBind10:}
\citep{le2018comprehensive} provides neural network predicted estimates of the relative binding affinities between all unique length-10 DNA sequences and each of two protein targets. For each target, the goal to identify DNA sequences that maximize the predicted binding affinity.

\textbf{UTR:}
\citet{sample2019human} introduced a CNN to predict the impact of a 5\textquoteright UTR sequence on the expression level of its corresponding gene. The goal is to identify length-50 DNA sequences that maximize the predicted expression level.

\textbf{RandomMLP/RandomRNN:} Following \citet{brookes2018design}, we design optimization problems with the goal find inputs that maximize the scalar output of randomly initialized fully-connected or recurrent neural networks.

\addtolength{\tabcolsep}{-4pt}
\begin{table}[t]
    \centering
    \begin{footnotesize}
    \begin{tabular}{l|cc|ccccc}
    \toprule

\multirow{2}{*}{Problem} &  Adap. &  \multirow{2}{*}{P3BO} & \multirow{2}{*}{MBO} &  DbAs &  Latent &  \multirow{2}{*}{Evo} & \multirow{2}{*}{SMW} \\
                      &  P3BO     &       &      &  VAE  &   MBO& & \\
\midrule
PdbIsing        &  \bf 7.0 &  5.6 &  5.2 &  3.8 &  3.0 &  2.3 &  1.1 \\
PfamHMM         &  5.8 &  5.0 &  2.4 & \bf 6.0 &  3.6 &  2.2 &  3.0 \\
ProteinDist.    &  \bf 6.8 &  6.2 &  4.8 &  3.0 &  4.1 &  2.1 &  1.0 \\
RandomMLP       &  \bf 7.0 &  5.9 &  4.5 &  3.6 &  3.9 &  2.0 &  1.0 \\
RandomRNN       &  6.2 &  \bf 6.7 &  5.1 &  1.2 &  2.9 &  3.7 &  2.2 \\
TfBind10        &  5.0 &  \bf 6.5 & \bf  6.5 &  3.0 &  2.0 &  1.0 &  4.0 \\
TfBind8         &  \bf 6.8 &  5.9 &  4.5 &  3.0 &  1.3 &  1.9 &  4.6 \\
UTR             &  \bf 7.0 &  6.0 &  2.0 &  4.0 &  1.0 &  5.0 &  3.0 \\
\bottomrule
\end{tabular}
\end{footnotesize}
    \caption{Ranking of methods in terms of the area under the \textit{Max reward} curve per optimization class (higher is better). Shown is the mean rank over all instances per optimization problem (\stabref{tab:app-problems}{1}). As also shown in \figref{fig:rank}, both \pbo and Adaptive-\pbo outperform or are comparable to all other baseline methods for sequence optimization.} 
    \label{tab:1}
    \vspace{-1em}
\end{table}
\addtolength{\tabcolsep}{4pt}

\vspace{1em}
\textbf{PfamHMM:} Pfam~\cite{el2018pfam} is a widely-used database of families of protein domain sequences. Each family consists of a small set of human-curated seed sequences, along with sequences that have been added automatically based on the likelihood under a profile hidden Markov model (HMM) fit using the seed sequences~\cite{finn2011hmmer}. We construct optimization problems for 24 diverse protein families. For each, we use the likelihood of the profile HMM as a black-box objective function. All optimization methods are provided with an initial dataset consisting of a random subset of the sequences in the family with a likelihood below the 50th percentile. This simulates practitioners' use of tools such as HMMer~\cite{finn2011hmmer} or HHblits~\cite{remmert2012hhblits} to find a set of evolutionarily-related sequences to inform sequence design.

\begin{figure}[t]
    \centering
    \includegraphics[width=.9\columnwidth]{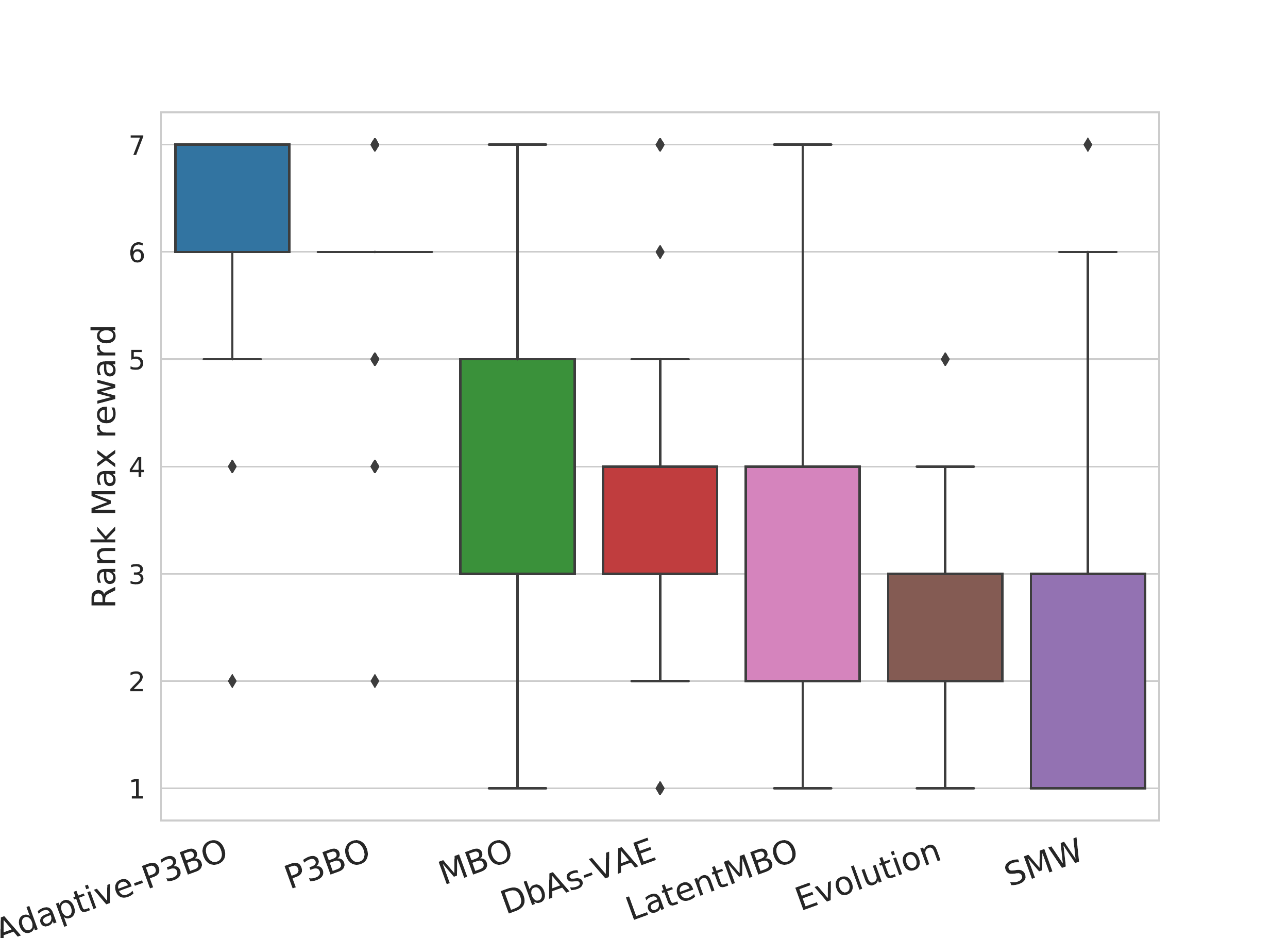}
    \caption{Ranking of methods in terms of the area under the \textit{Max reward} curve per optimization problem (higher is better).  Shown is the distribution of ranks over all instances per optimization problem (\stabref{tab:app-problems}{1}). The box of \pbo is flat since both the 25th and 75th percentile over all 105 optimization problems is 6.}
    \label{fig:rank}
    \vspace{-1em}
\end{figure}

\textbf{ProteinDistance:} \citet{bileschi2019using} introduced a CNN for protein domain classification that yields informative 1100-dimensional embeddings of protein domains. The ProteinDistance problem tasks methods to find sequences with high cosine similarity in the embedding space of this model to representative sequences from Pfam families. 

\textbf{PDBIsing:} 
As introduced in~\citet{Angermueller2020RLDesign}, the goal of this problem is to find protein sequences that maximize the energy of an Ising model parameterized by a protein structure from the Protein Data Bank~\cite{berman2003protein} (PDB). We reweight the objective function to place more emphasis on long-range pairwise interactions between amino acids. Optimization methods are provided with an initial dataset of mutants of the sequence that the PDB structure is based on.

\section{Experiments}
We next analyze the performance of a set of competitive optimization methods on the benchmark problems described the previous section.
\vspace{-1em}
\begin{figure*}[tb]
\includegraphics[width=\textwidth]{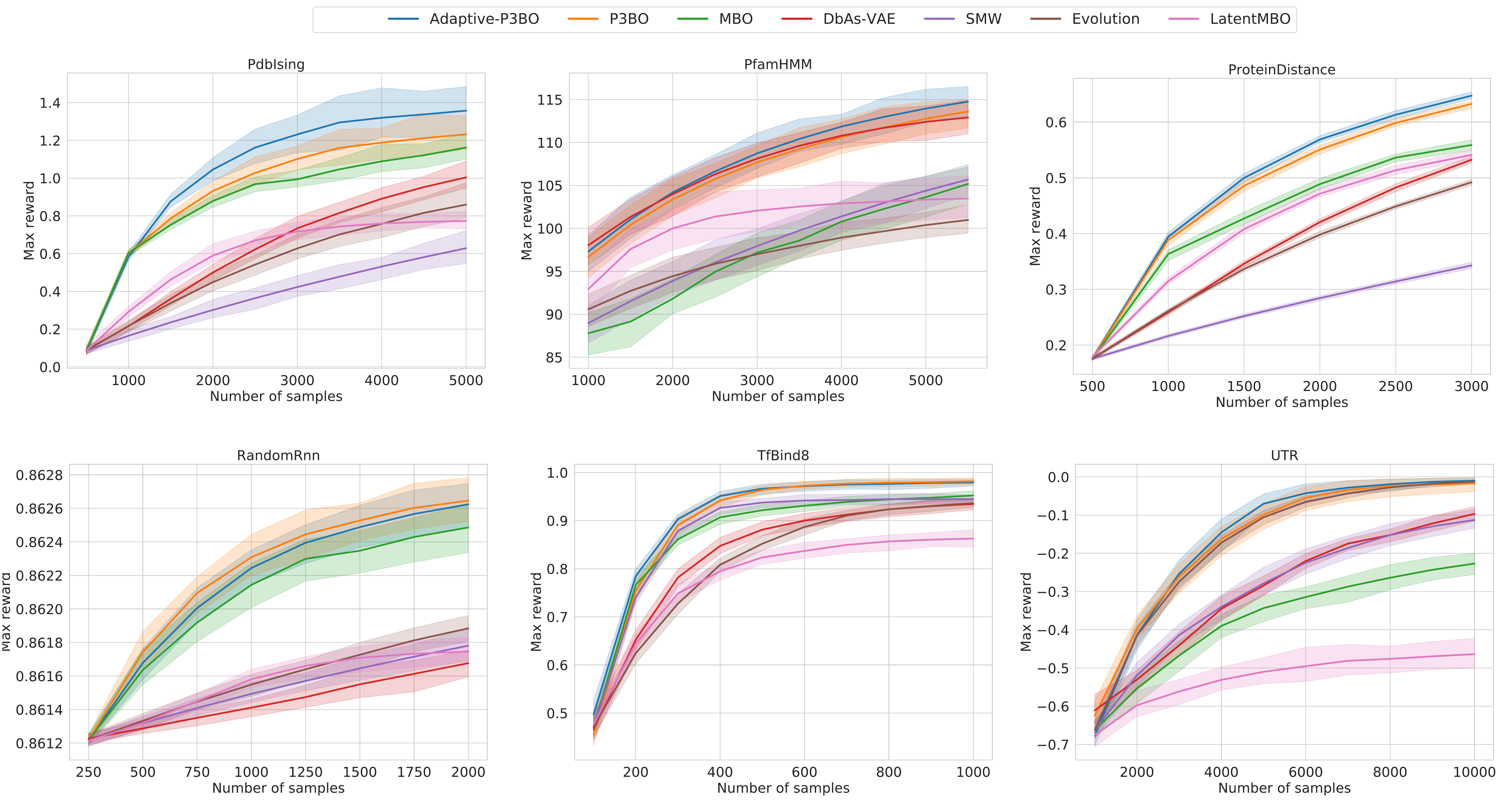}
\caption{Performances curves for different problem classes. Lines show the average over all instances per problem class.  The shaded area corresponds to the 95\% bootstrap confidence interval. The synergistic effect of the ensembling by \pbo is noteworthy: not only does it match the performance of the best ensemble member, but it outperforms it. Adaptive \pbo provides a further improvements in most cases.  Due to space limitations, results for RandomMLP and TfBind10 are shown in~\sfigref{fig:perf_other_problems}{2}, and results for additional baselines in~\sfigref{fig:perf_baselines}{3}.}
\label{fig:perf-main-text}
\vspace{-1em}
\end{figure*}

\subsection{Baseline Optimization Methods}
\vspace{-.5em}
\label{sec:exp-methods}
We consider the following methods that have been designed for batched black-box optimization of biological sequences.

\textbf{SingleMutantWalker (SMW)} simulates site-saturation mutagenesis, where all  single-mutation neighbors of the best sequence seen so far are proposed in a given optimization round~\cite{wu2019machine}. 

\textbf{Evolution} performs directed evolutionary search by selecting the top $k$ sequences, recombining them, and mutating them~\cite{unter1958global, brindle1980genetic}.

\textbf{Cross-Entropy Method} fits a generative model to maximize the likelihood of high-quality sequences and samples the next batch of sequences from this model. \textbf{DbAs-VAE} uses the same example weighting scheme and generative model, a fully-connected variational autoencoder~\cite{vae}, as in~\citep{brookes2018design}. Supplementary material also considers \textbf{FBGAN}~\citep{gupta2018feedback}, which uses a generative adversarial network.

\textbf{Model-Based Optimization (MBO)} automatically tunes the hyper-parameters of diverse candidate regressor models as described in~\cite{Angermueller2020RLDesign}. All models with cross-validation performance above a predefined threshold are ensembled, yielding a predicted mean and variance for each seqeunce. These are converted into an acquisition function (\eg, expected improvement), which is optimized with regularized evolution~\citep{real2019regularized} to yield the next batch of sequences.

\textbf{Latent-Space MBO (LatentMBO):} We adapt the method of \cite{gomez2018automatic} to batched optimization. Model-based optimization is performed in the continuous latent space of a generative model that is trained jointly with a regressor model, which predicts $\fx$ given the latent embedding of $\mathbf{x}$. We use the cross-entropy method for optimizing the regressor.

\subsection{Population-Based Optimization}
\label{sec:exp-pbo}
\vspace{-.5em}
We used a population with $N=15$ constituent algorithms belonging to the following classes: \textbf{SMW}, \textbf{MBO} with various regressors, acquisition functions, and mutation rates of the evolution acquisition function optimizer; \textbf{DbAs} with different generative models (VAE or LSTM) and quantile thresholds; and \textbf{Evolution} with varying mutation and crossover probabilities. See \ssecref{sec:app-pbo}{B.7} for more details.

\begin{figure*}[t]
\centering
\includegraphics[width=1.0\textwidth]{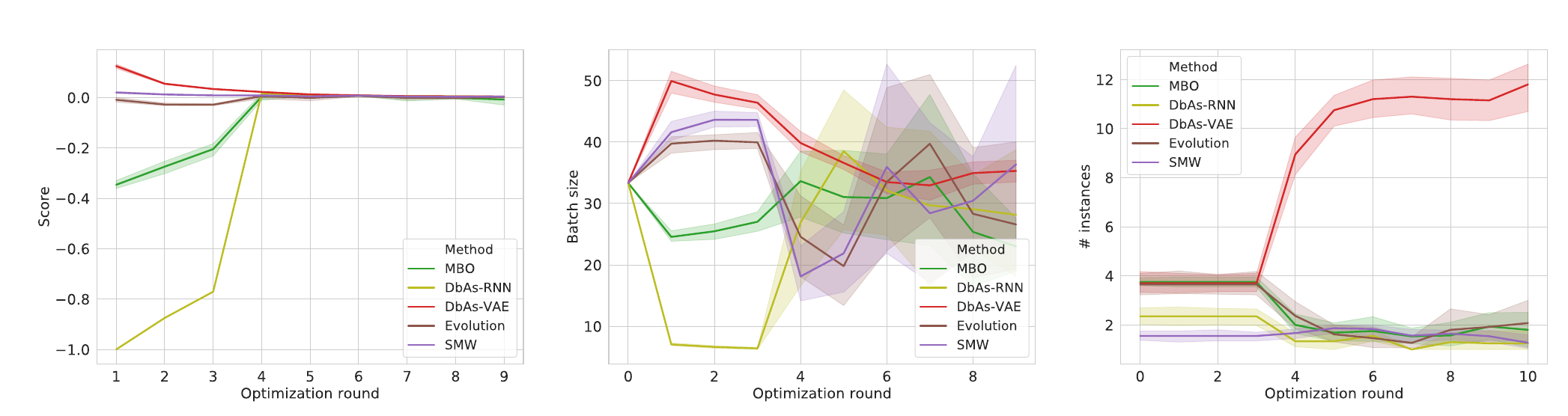}
\caption{Insights into Adaptive-\pbo applied to PfamHMM target PF16186. Shown are the credit score (left), the number of sequences sampled (middle), and the number of instances (right) per algorithm class over time. Since DbAs-VAE has the highest credit score (relative improvement) in early rounds, more sequences are sampled from DbAs-VAE (middle), and Adaptive-\pbo increases the number of DbAs-VAE instances from 4 to 11 (the total population size is 15). The adaptation starts after three warm-up rounds used to reliably estimate the credit score of algorithms. See \sfigref{fig:pbo_members_utr}{4} for another example.}
\label{fig:pbo_members_pfam}
\end{figure*}
\begin{figure*}[t]
\centering
\includegraphics[width=.5\textwidth]{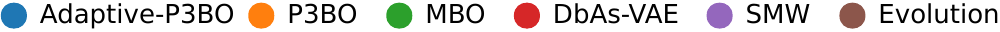}

\subfigure[Batch 1]{\includegraphics[width=.24\textwidth]{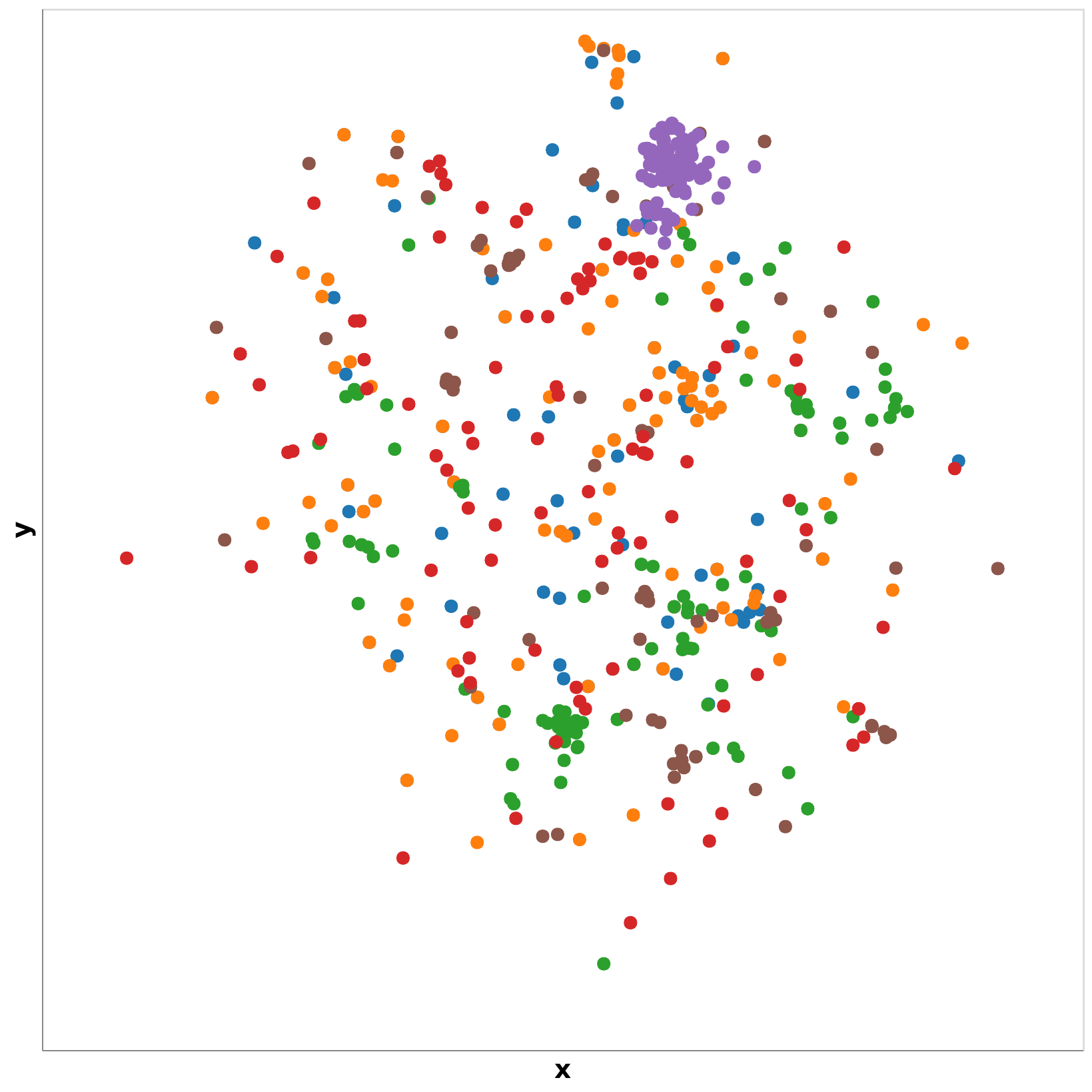}\label{fig:tsne-b=1}}
\subfigure[Batch 5]{\includegraphics[width=.24\textwidth]{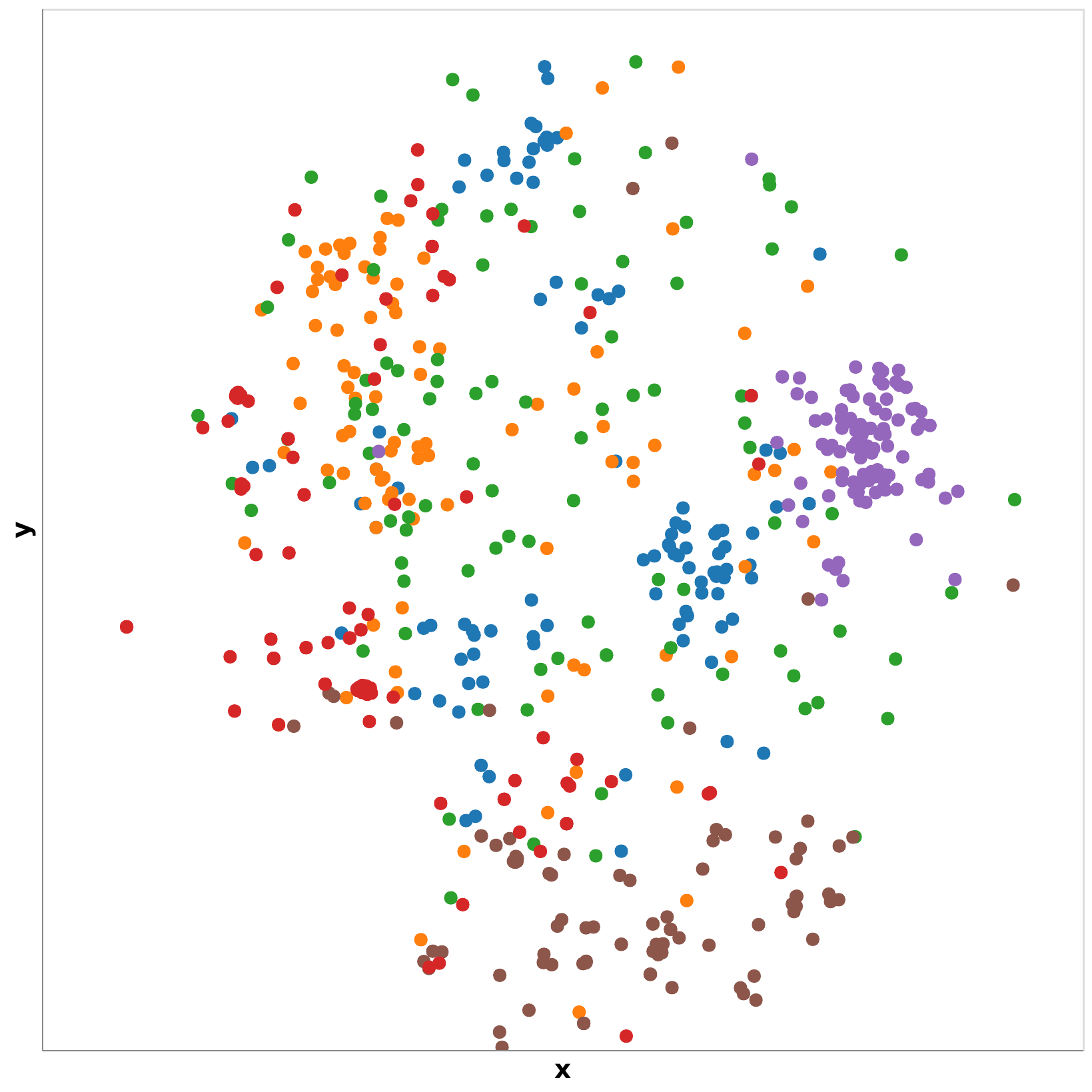}\label{fig:tsne-b=5}}
\subfigure[Batch 9]{\includegraphics[width=.24\textwidth]{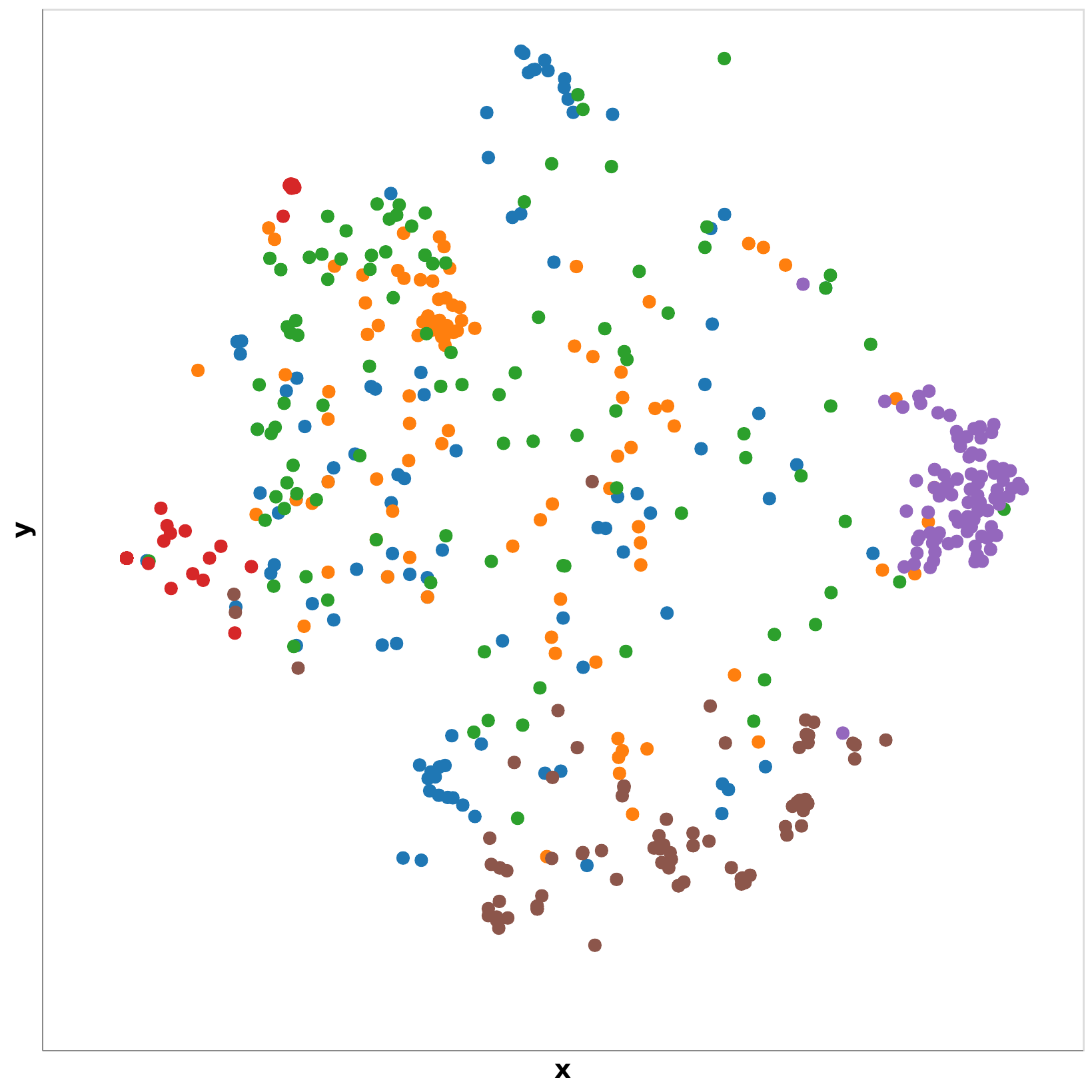}\label{fig:tsne-b=9}}
\subfigure[Rewards in 75th percentile]{\includegraphics[width=.24\textwidth]{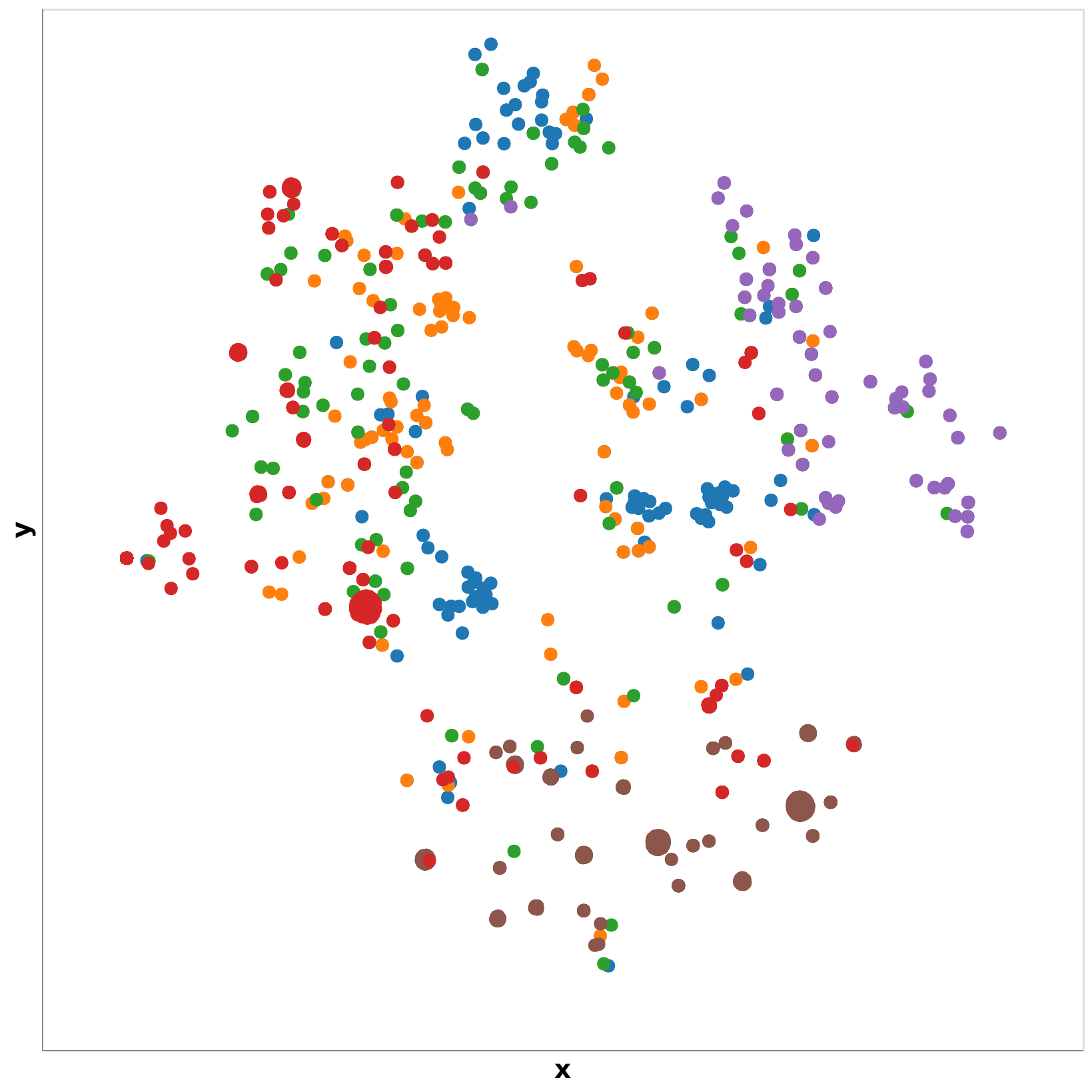}\label{fig:tsne-threshold}}
\vspace{-1em}
\caption{$t$-SNE plots of sequences found by methods in different rounds (a-c), and of all sequences with a reward greater than the 75th percentile of rewards found by all methods (d). Results are for TfBind8, targetet CRX\_R90W\_R1. SMW, Evolution, and DbAs-VAE increasingly focus on a single  search region, whereas MBO, \pbo and Adaptive-\pbo maintain a diversity of sequences within each batch (Fig. a-c). When showing only sequences that achieved a relatively high reward in \figref{fig:tsne-threshold}, we see that \pbo and Adaptive-\pbo still find diverse clusters of high-reward sequences. In fact, \pbo and Adaptive-\pbo find sequences in nearly all clusters identified by baseline methods.}
\label{fig:tsne}
\vspace{-1em}
\end{figure*}

We sampled the initial population of algorithms randomly, ensuring that \pbo includes at least one instance per class to promote diversity, and at most four MBO instances to reduce computational costs. We used a decay factor of $\gamma=0.25$ for computing credit scores, and a softmax temperature of $\tau=1.0$ for computing selection probabilities. 

We compare \pbo to Adaptive-\pbo, which adapts population members as described in Section~\ref{sec:adaptive-pbo}. For Adaptive-\pbo, we use a quantile cutoff of $q=0.5$ for selecting the pool of survivors $\mathcal{S}$, a recombination rate of $0.1$, and a mutation rate of $0.5$. Experimentally, we observed that the performance of Adaptive-\pbo remained robust to any of these hyper-parameters. 

\subsection{Evaluation of Sample-Efficiency and Robustness}
\vspace{-.5em}
We evaluate sample-efficiency for a particular optimization problem by comparing the cumulative maximum of $\fx$ (\textit{Max reward}) depending on the number of samples proposed. We further use the area under the \textit{Max reward} curve to summarize sample-efficiency in a single number and comparing methods across optimization problems. We repeat each experiment 20 times with different random seeds.

\figref{fig:perf-main-text} illustrates the mean optimization trajectories of \pbo, Adaptive-\pbo, and baselines over 6 different problem classes. \pbo and Adaptive-\pbo systematically find high-reward sequences faster than any baseline method, which shows that population-based optimization can increase both sample-efficiency and robustness. Adapting the population of optimizers (Adaptive-\pbo) yields further performance improvements. Non-population-based methods have inconsistent performance: there is no best non-population-based method across problems. Optimization trajectories for additional problem classes and baseline methods are shown in \sfigref{fig:perf_other_problems}{2} and \sfigref{fig:perf_baselines}{3}, respectively.

Table~\ref{tab:1} and \figref{fig:rank} summarize results, which report the average ranking of different optimization methods across all 105 optimization problems. \pbo and \apbo are systematically more sample-efficient than any baseline method.

\subsection{Evaluation of Diversity}

\begin{figure}[!b!]
    \centering
    \includegraphics[width=\columnwidth]{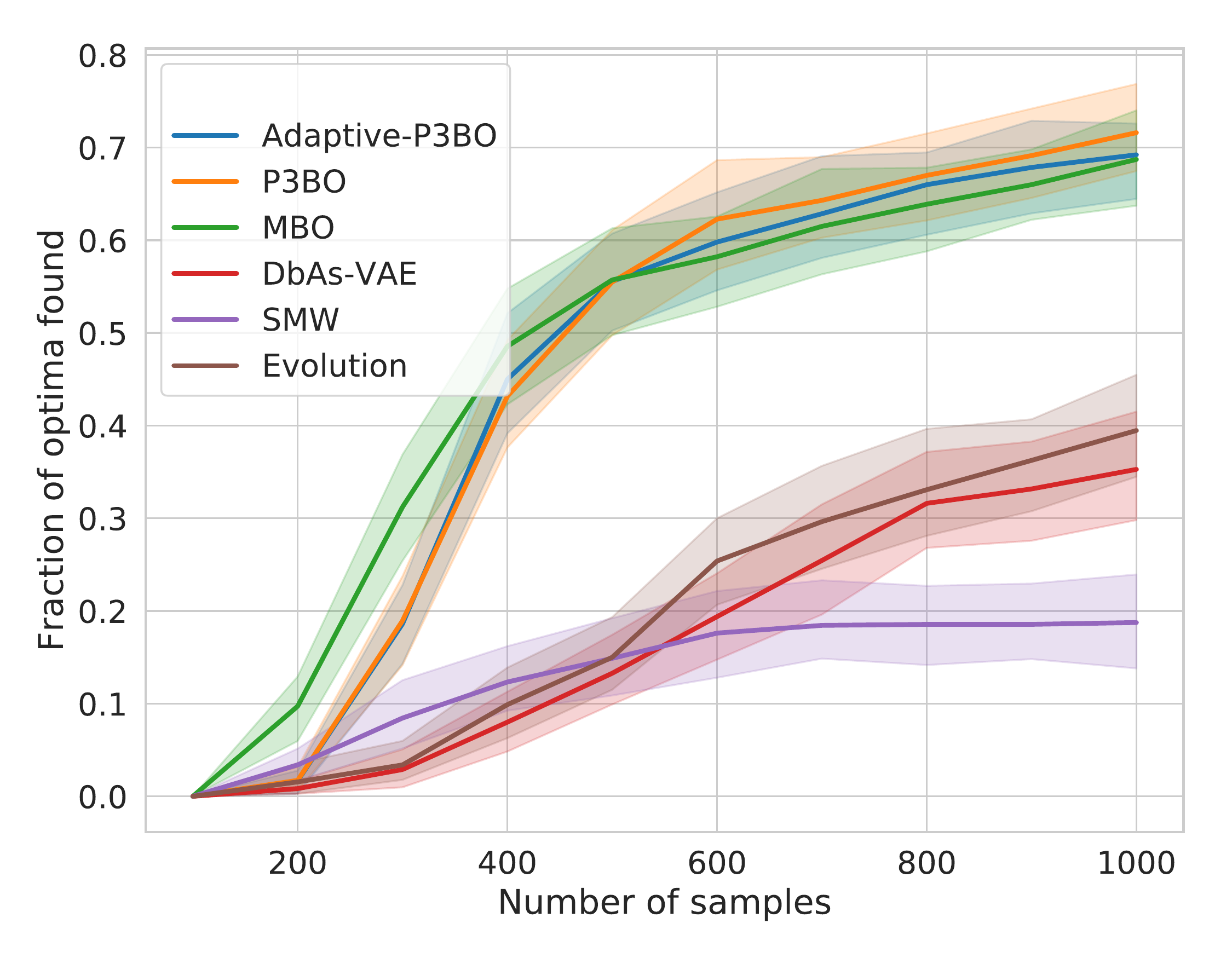}
    \caption{Fraction of optima found by each method averaged over all TfBind8 targets.}
    \label{fig:fraction}
    \vspace{-.8em}
\end{figure}

We evaluate the ability of methods to find distinct sequences of high reward using the metrics summarized in \ssecref{sec:app-diversity}{C}. 

\figref{fig:fraction} reports one of these metrics---the fraction of optima found by each method for TfBind8. \pbo and Adaptive-\pbo find more optima than the best individual method (MBO), and considerably more optima than all remaining methods. \sfigref{fig:diversity}{1} further shows the mean distance of sequences in proposed batches, and the number of distinct clusters of sequences with a high reward that were found by methods. These results confirm that \pbo and Adaptive-\pbo find diverse sequences with a high reward across optimization problems.

\figref{fig:tsne} uses $t$-SNE to illustrate the diversity of sequences obtained over time and at the end of optimization for each method. \pbo and Adaptive-\pbo not only find more diverse sequences per batch during the optimization, but also obtain diverse \emph{and} high-reward sequences at the end of the optimization, fulfilling a crucial requirement for real-world sequence validation. 

\subsection{Ablation Experiments}
We perform ablation experiments to better understand the behavior of \pbo and \apbo. 

\figref{fig:pbo_members_pfam} illustrates how the scoring of algorithms affects the number of sequences sampled from each algorithm class and the number of instances per class for one PfamHMM problem. \apbo correctly identifies DbAs-VAE as the best performing algorithm (see \figref{fig:perf-main-text}, PfamHMM), and hence samples more sequences from DbAs-VAE and enriches for the number of DbAs-VAE instances in the population. We observe the same behavior on other optimization problems (\sfigref{fig:pbo_members_utr}{4}).

\sfigref{fig:abl_01}{5(a)} analyzes the performance of \pbo when removing individual algorithm classes from its population. As expected, the performance of P3BO is most sensitive to the best performing algorithm in its population. However, since the performance of algorithms varies across problems, removing individual algorithms results in a performance drop on some problems but not on others. \sfigref{fig:abl}{5} also shows that sharing data between optimization methods enables poorly performing algorithms to learn from the sequences found by well performing algorithms and to find high-quality sequences faster, which results in an overall performance gain of P3BO. This is confirmed visually by \sfigref{fig:tsne_data_sharing}{6}, which shows the $t$-SNE visualization of sequences proposed by individual algorithms over time with and without data sharing.

\sfigref{fig:adaptive_pbo_bad_init}{6} highlights that  \apbo recovers faster than \pbo when starting with a poorly initialized population of algorithms, which it can improve online by evolutionary search as described in Section~\ref{sec:adaptive-pbo}.

Finally, we investigate the sensitivity of \pbo to the choice of the scoring function (\sfigref{fig:score_random,fig:score_rank}{8, 9}), the temperature $\tau$ of the softmax scoring function (\sfigref{fig:score_temp}{10}), the population size (\sfigref{fig:psize}{11}), and the batch size of the optimization problem (\sfigref{fig:perf_vs_batch_size}{12}).

\vspace{-.5em}
\section{Conclusion}
\vspace{-.5em}
We show on extensive novel \emph{in-silico} benchmarks that standard ML approaches to sequence design lack robustness across optimization problems. By ensembling over algorithms, \pbo increases sample-efficiency, robustness, and the ability to discover diverse optima. Online adaptation of the population of algorithms (Adaptive-\pbo) further improves performance. Efficient ways of selecting the initial population and enabling transfer learning across optimization problems remain crucial open questions. Finally, although \pbo is motivated by biological sequence design, \pbo is also applicable to non-biological black-box optimization problems.
\clearpage
\bibliography{sources}

\begin{thebibliography}{43}
\providecommand{\natexlab}[1]{#1}
\providecommand{\url}[1]{\texttt{#1}}
\expandafter\ifx\csname urlstyle\endcsname\relax
  \providecommand{\doi}[1]{doi: #1}\else
  \providecommand{\doi}{doi: \begingroup \urlstyle{rm}\Url}\fi

\bibitem[Angermueller et~al.(2020)Angermueller, Dohan, Belanger, Deshpande,
  Murphy, and Colwell]{Angermueller2020RLDesign}
Angermueller, C., Dohan, D., Belanger, D., Deshpande, R., Murphy, K., and
  Colwell, L.
\newblock Model-based reinforcement learning for biological sequence design.
\newblock In \emph{ICLR}, 2020.

\bibitem[Arnold(1998)]{arnold1998design}
Arnold, F.~H.
\newblock Design by directed evolution.
\newblock \emph{Accounts of chemical research}, 31\penalty0 (3), 1998.

\bibitem[Barrera et~al.(2016)Barrera, Vedenko, Kurland, Rogers, Gisselbrecht,
  Rossin, Woodard, Mariani, Kock, Inukai, Siggers, Shokri, Gord{\^a}n, Sahni,
  Cotsapas, Hao, Yi, Kellis, Daly, Vidal, Hill, and Bulyk]{barrera2016survey}
Barrera, L.~A., Vedenko, A., Kurland, J.~V., Rogers, J.~M., Gisselbrecht,
  S.~S., Rossin, E.~J., Woodard, J., Mariani, L., Kock, K.~H., Inukai, S.,
  Siggers, T., Shokri, L., Gord{\^a}n, R., Sahni, N., Cotsapas, C., Hao, T.,
  Yi, S., Kellis, M., Daly, M.~J., Vidal, M., Hill, D.~E., and Bulyk, M.~L.
\newblock Survey of variation in human transcription factors reveals prevalent
  {DNA} binding changes.
\newblock \emph{Science}, 2016.

\bibitem[Berman et~al.(2003)Berman, Bourne, Westbrook, and
  Zardecki]{berman2003protein}
Berman, H.~M., Bourne, P.~E., Westbrook, J., and Zardecki, C.
\newblock The protein data bank.
\newblock In \emph{Protein Structure}. CRC Press, 2003.

\bibitem[Bileschi et~al.(2019)Bileschi, Belanger, Bryant, Sanderson, Carter,
  Sculley, DePristo, and Colwell]{bileschi2019using}
Bileschi, M.~L., Belanger, D., Bryant, D.~H., Sanderson, T., Carter, B.,
  Sculley, D., DePristo, M.~A., and Colwell, L.~J.
\newblock Using deep learning to annotate the protein universe.
\newblock \emph{bioRxiv}, 2019.

\bibitem[Biswas et~al.(2020)Biswas, Khimulya, Alley, Esvelt, and
  Church]{biswas2020low}
Biswas, S., Khimulya, G., Alley, E.~C., Esvelt, K.~M., and Church, G.~M.
\newblock Low-n protein engineering with data-efficient deep learning.
\newblock \emph{bioRxiv}, 2020.

\bibitem[Brindle(1980)]{brindle1980genetic}
Brindle, A.
\newblock Genetic algorithms for function optimization.
\newblock 1980.

\bibitem[Brookes \& Listgarten(2018)Brookes and Listgarten]{brookes2018design}
Brookes, D.~H. and Listgarten, J.
\newblock Design by adaptive sampling.
\newblock \emph{arXiv preprint arXiv:1810.03714}, 2018.

\bibitem[Brookes et~al.(2019)Brookes, Busia, Fannjiang, Murphy, and
  Listgarten]{brookes2019view}
Brookes, D.~H., Busia, A., Fannjiang, C., Murphy, K., and Listgarten, J.
\newblock A view of estimation of distribution algorithms through the lens of
  expectation-maximization, 2019.

\bibitem[Burke et~al.(2013)Burke, Gendreau, Hyde, Kendall, Ochoa, {\"O}zcan,
  and Qu]{burke2013hyper}
Burke, E.~K., Gendreau, M., Hyde, M., Kendall, G., Ochoa, G., {\"O}zcan, E.,
  and Qu, R.
\newblock Hyper-heuristics: A survey of the state of the art.
\newblock \emph{Journal of the Operational Research Society}, 64\penalty0
  (12):\penalty0 1695--1724, 2013.

\bibitem[De~Boer et~al.(2005)De~Boer, Kroese, Mannor, and
  Rubinstein]{de2005tutorial}
De~Boer, P.-T., Kroese, D.~P., Mannor, S., and Rubinstein, R.~Y.
\newblock A tutorial on the cross-entropy method.
\newblock \emph{Annals of operations research}, 134\penalty0 (1), 2005.

\bibitem[de~Jongh et~al.(2019)de~Jongh, van Dijk, Julsing, Schaap, and
  de~Ridder]{de2019designing}
de~Jongh, R.~P., van Dijk, A.~D., Julsing, M.~K., Schaap, P.~J., and de~Ridder,
  D.
\newblock Designing eukaryotic gene expression regulation using machine
  learning.
\newblock \emph{Trends in biotechnology}, 2019.

\bibitem[Du \& Li(2008)Du and Li]{du2008multi}
Du, W. and Li, B.
\newblock Multi-strategy ensemble particle swarm optimization for dynamic
  optimization.
\newblock \emph{Information sciences}, 178\penalty0 (15):\penalty0 3096--3109,
  2008.

\bibitem[El-Gebali et~al.(2018)El-Gebali, Mistry, Bateman, Eddy, Luciani,
  Potter, Qureshi, Richardson, Salazar, Smart, et~al.]{el2018pfam}
El-Gebali, S., Mistry, J., Bateman, A., Eddy, S.~R., Luciani, A., Potter,
  S.~C., Qureshi, M., Richardson, L.~J., Salazar, G.~A., Smart, A., et~al.
\newblock The {Pfam} protein families database in 2019.
\newblock \emph{Nucleic acids research}, 47\penalty0 (D1), 2018.

\bibitem[Fialho et~al.(2010)Fialho, Schoenauer, and Sebag]{fialho2010toward}
Fialho, {\'A}., Schoenauer, M., and Sebag, M.
\newblock Toward comparison-based adaptive operator selection.
\newblock In \emph{Proceedings of the 12th annual conference on Genetic and
  evolutionary computation}, pp.\  767--774, 2010.

\bibitem[Finn et~al.(2011)Finn, Clements, and Eddy]{finn2011hmmer}
Finn, R.~D., Clements, J., and Eddy, S.~R.
\newblock {HMMER} web server: interactive sequence similarity searching.
\newblock \emph{Nucleic acids research}, 39\penalty0 (suppl\_2), 2011.

\bibitem[G{\'o}mez-Bombarelli et~al.(2018)G{\'o}mez-Bombarelli, Wei, Duvenaud,
  Hern{\'a}ndez-Lobato, S{\'a}nchez-Lengeling, Sheberla, Aguilera-Iparraguirre,
  Hirzel, Adams, and Aspuru-Guzik]{gomez2018automatic}
G{\'o}mez-Bombarelli, R., Wei, J.~N., Duvenaud, D., Hern{\'a}ndez-Lobato,
  J.~M., S{\'a}nchez-Lengeling, B., Sheberla, D., Aguilera-Iparraguirre, J.,
  Hirzel, T.~D., Adams, R.~P., and Aspuru-Guzik, A.
\newblock Automatic chemical design using a data-driven continuous
  representation of molecules.
\newblock \emph{ACS central science}, 4\penalty0 (2), 2018.

\bibitem[Gupta \& Zou(2018)Gupta and Zou]{gupta2018feedback}
Gupta, A. and Zou, J.
\newblock Feedback {GAN (FBGAN)} for {DNA}: A novel feedback-loop architecture
  for optimizing protein functions.
\newblock \emph{arXiv preprint arXiv:1804.01694}, 2018.

\bibitem[Hashimoto et~al.(2018)Hashimoto, Yadlowsky, and
  Duchi]{hashimoto2018derivative}
Hashimoto, T.~B., Yadlowsky, S., and Duchi, J.~C.
\newblock Derivative free optimization via repeated classification, 2018.

\bibitem[Jaderberg et~al.(2017)Jaderberg, Dalibard, Osindero, Czarnecki,
  Donahue, Razavi, Vinyals, Green, Dunning, Simonyan,
  et~al.]{jaderberg2017population}
Jaderberg, M., Dalibard, V., Osindero, S., Czarnecki, W.~M., Donahue, J.,
  Razavi, A., Vinyals, O., Green, T., Dunning, I., Simonyan, K., et~al.
\newblock Population based training of neural networks.
\newblock \emph{arXiv preprint arXiv:1711.09846}, 2017.

\bibitem[Khadka \& Tumer(2018)Khadka and Tumer]{khadka2018evolution}
Khadka, S. and Tumer, K.
\newblock Evolution-guided policy gradient in reinforcement learning.
\newblock In \emph{Advances in Neural Information Processing Systems}, pp.\
  1188--1200, 2018.

\bibitem[Kingma \& Welling(2014)Kingma and Welling]{vae}
Kingma, D.~P. and Welling, M.
\newblock Auto-encoding variational {B}ayes.
\newblock In \emph{2nd International Conference on Learning Representations,
  {ICLR} 2014, Banff, AB, Canada, April 14-16, 2014, Conference Track
  Proceedings}, 2014.

\bibitem[Le et~al.(2018)Le, Shimko, Aditham, Keys, Longwell, Orenstein, and
  Fordyce]{le2018comprehensive}
Le, D.~D., Shimko, T.~C., Aditham, A.~K., Keys, A.~M., Longwell, S.~A.,
  Orenstein, Y., and Fordyce, P.~M.
\newblock Comprehensive, high-resolution binding energy landscapes reveal
  context dependencies of transcription factor binding.
\newblock \emph{Proceedings of the National Academy of Sciences}, 115\penalty0
  (16):\penalty0 E3702--E3711, 2018.

\bibitem[Lehman \& Stanley(2008)Lehman and Stanley]{lehman2008exploiting}
Lehman, J. and Stanley, K.~O.
\newblock Exploiting open-endedness to solve problems through the search for
  novelty.
\newblock In \emph{ALIFE}, pp.\  329--336, 2008.

\bibitem[Leyton-Brown et~al.(2003)Leyton-Brown, Nudelman, Andrew, McFadden, and
  Shoham]{leyton-brown_portfolio_2003}
Leyton-Brown, K., Nudelman, E., Andrew, G., McFadden, J., and Shoham, Y.
\newblock A portfolio approach to algorithm selection.
\newblock In \emph{IJCAI}, volume~3, pp.\  1542--1543, 2003.

\bibitem[Li et~al.(2013)Li, Fialho, Kwong, and Zhang]{li2013adaptive}
Li, K., Fialho, A., Kwong, S., and Zhang, Q.
\newblock Adaptive operator selection with bandits for a multiobjective
  evolutionary algorithm based on decomposition.
\newblock \emph{IEEE Transactions on Evolutionary Computation}, 18\penalty0
  (1):\penalty0 114--130, 2013.

\bibitem[Littlestone et~al.(1989)Littlestone, Warmuth,
  et~al.]{littlestone1989weighted}
Littlestone, N., Warmuth, M.~K., et~al.
\newblock \emph{The weighted majority algorithm}.
\newblock University of California, Santa Cruz, Computer Research Laboratory,
  1989.

\bibitem[Liu et~al.(2019)Liu, Zeng, Mueller, Carter, Wang, Schilz, Horny,
  Birnbaum, Ewert, and Gifford]{liu2019antibody}
Liu, G., Zeng, H., Mueller, J., Carter, B., Wang, Z., Schilz, J., Horny, G.,
  Birnbaum, M.~E., Ewert, S., and Gifford, D.~K.
\newblock Antibody complementarity determining region design using
  high-capacity machine learning.
\newblock \emph{bioRxiv}, 2019.

\bibitem[Maturana et~al.(2009)Maturana, Fialho, Saubion, Schoenauer, and
  Sebag]{maturana2009extreme}
Maturana, J., Fialho, {\'A}., Saubion, F., Schoenauer, M., and Sebag, M.
\newblock Extreme compass and dynamic multi-armed bandits for adaptive operator
  selection.
\newblock In \emph{2009 IEEE Congress on Evolutionary Computation}, pp.\
  365--372. IEEE, 2009.

\bibitem[Miller et~al.(1995)Miller, Goldberg, et~al.]{miller1995genetic}
Miller, B.~L., Goldberg, D.~E., et~al.
\newblock Genetic algorithms, tournament selection, and the effects of noise.
\newblock \emph{Complex systems}, 9\penalty0 (3):\penalty0 193--212, 1995.

\bibitem[Mockus et~al.(2014)Mockus, Tiesis, and
  Zilinskas]{mockus1978application}
Mockus, J., Tiesis, V., and Zilinskas, A.
\newblock The application of {B}ayesian methods for seeking the extremum.
\newblock \emph{Towards Global Optimization}, 2:\penalty0 117--129, 09 2014.

\bibitem[Neil et~al.(2018)Neil, Segler, Guasch, Ahmed, Plumbley, Sellwood, and
  Brown]{neil2018exploring}
Neil, D., Segler, M., Guasch, L., Ahmed, M., Plumbley, D., Sellwood, M., and
  Brown, N.
\newblock Exploring deep recurrent models with reinforcement learning for
  molecule design.
\newblock In \emph{International Conference on Learning Representations
  Workshop}, 2018.

\bibitem[Pourchot \& Sigaud(2018)Pourchot and Sigaud]{pourchot2018cem}
Pourchot, A. and Sigaud, O.
\newblock Cem-rl: Combining evolutionary and gradient-based methods for policy
  search.
\newblock \emph{arXiv preprint arXiv:1810.01222}, 2018.

\bibitem[Real et~al.(2019)Real, Aggarwal, Huang, and Le]{real2019regularized}
Real, E., Aggarwal, A., Huang, Y., and Le, Q.~V.
\newblock Regularized evolution for image classifier architecture search.
\newblock In \emph{Proceedings of the AAAI Conference on Artificial
  Intelligence}, volume~33, 2019.

\bibitem[Remmert et~al.(2012)Remmert, Biegert, Hauser, and
  S{\"o}ding]{remmert2012hhblits}
Remmert, M., Biegert, A., Hauser, A., and S{\"o}ding, J.
\newblock {HHblits}: lightning-fast iterative protein sequence searching by
  {HMM-HMM} alignment.
\newblock \emph{Nature methods}, 9\penalty0 (2), 2012.

\bibitem[Sample et~al.(2019)Sample, Wang, Reid, Presnyak, McFadyen, Morris, and
  Seelig]{sample2019human}
Sample, P.~J., Wang, B., Reid, D.~W., Presnyak, V., McFadyen, I.~J., Morris,
  D.~R., and Seelig, G.
\newblock Human 5 {UTR} design and variant effect prediction from a massively
  parallel translation assay.
\newblock \emph{Nature biotechnology}, 37\penalty0 (7), 2019.

\bibitem[Tang et~al.(2014)Tang, Peng, Chen, and
  Yao]{tang_population-based_2014}
Tang, K., Peng, F., Chen, G., and Yao, X.
\newblock Population-based {Algorithm} {Portfolios} with automated constituent
  algorithms selection.
\newblock \emph{Information Sciences}, 279:\penalty0 94--104, 2014.

\bibitem[unter Rudolph(1958)]{unter1958global}
unter Rudolph, G.
\newblock Global optimization by means of distributed evolution strategies.
\newblock \emph{Schwefel and M anner}, 2035, 1958.

\bibitem[Vinyals et~al.(2019)Vinyals, Babuschkin, Czarnecki, Mathieu, Dudzik,
  Chung, Choi, Powell, Ewalds, Georgiev, et~al.]{vinyals2019grandmaster}
Vinyals, O., Babuschkin, I., Czarnecki, W.~M., Mathieu, M., Dudzik, A., Chung,
  J., Choi, D.~H., Powell, R., Ewalds, T., Georgiev, P., et~al.
\newblock Grandmaster level in starcraft ii using multi-agent reinforcement
  learning.
\newblock \emph{Nature}, 575\penalty0 (7782):\penalty0 350--354, 2019.

\bibitem[Wang et~al.(2019)Wang, Wang, Liu, and Wang]{wang2019synthetic}
Wang, Y., Wang, H., Liu, L., and Wang, X.
\newblock Synthetic promoter design in escherichia coli based on generative
  adversarial network.
\newblock \emph{BioRxiv}, 2019.

\bibitem[Wu et~al.(2019{\natexlab{a}})Wu, Mallipeddi, and
  Suganthan]{wu2019ensemble}
Wu, G., Mallipeddi, R., and Suganthan, P.~N.
\newblock Ensemble strategies for population-based optimization algorithms--a
  survey.
\newblock \emph{Swarm and evolutionary computation}, 44, 2019{\natexlab{a}}.

\bibitem[Wu et~al.(2019{\natexlab{b}})Wu, Kan, Lewis, Wittmann, and
  Arnold]{wu2019machine}
Wu, Z., Kan, S.~J., Lewis, R.~D., Wittmann, B.~J., and Arnold, F.~H.
\newblock Machine learning-assisted directed protein evolution with
  combinatorial libraries.
\newblock \emph{Proceedings of the National Academy of Sciences}, 116\penalty0
  (18), 2019{\natexlab{b}}.

\bibitem[Yang et~al.(2019)Yang, Wu, and Arnold]{yang2019machine}
Yang, K.~K., Wu, Z., and Arnold, F.~H.
\newblock Machine-learning-guided directed evolution for protein engineering.
\newblock \emph{Nature methods}, 2019.

\end{thebibliography}


\begin{thebibliography}{19}
\providecommand{\natexlab}[1]{#1}
\providecommand{\url}[1]{\texttt{#1}}
\expandafter\ifx\csname urlstyle\endcsname\relax
  \providecommand{\doi}[1]{doi: #1}\else
  \providecommand{\doi}{doi: \begingroup \urlstyle{rm}\Url}\fi

\bibitem[Angermueller et~al.(2020)Angermueller, Dohan, Belanger, Deshpande,
  Murphy, and Colwell]{Angermueller2020RLDesign}
Angermueller, C., Dohan, D., Belanger, D., Deshpande, R., Murphy, K., and
  Colwell, L.
\newblock Model-based reinforcement learning for biological sequence design.
\newblock In \emph{ICLR}, 2020.

\bibitem[Berman et~al.(2003)Berman, Bourne, Westbrook, and
  Zardecki]{berman2003protein}
Berman, H.~M., Bourne, P.~E., Westbrook, J., and Zardecki, C.
\newblock The protein data bank.
\newblock In \emph{Protein Structure}. CRC Press, 2003.

\bibitem[Bileschi et~al.(2019)Bileschi, Belanger, Bryant, Sanderson, Carter,
  Sculley, DePristo, and Colwell]{bileschi2019using}
Bileschi, M.~L., Belanger, D., Bryant, D.~H., Sanderson, T., Carter, B.,
  Sculley, D., DePristo, M.~A., and Colwell, L.~J.
\newblock Using deep learning to annotate the protein universe.
\newblock \emph{bioRxiv}, 2019.

\bibitem[Brookes \& Listgarten(2018)Brookes and Listgarten]{brookes2018design}
Brookes, D.~H. and Listgarten, J.
\newblock Design by adaptive sampling.
\newblock \emph{arXiv preprint arXiv:1810.03714}, 2018.

\bibitem[Brookes et~al.(2019)Brookes, Park, and
  Listgarten]{brookes2019conditioning}
Brookes, D.~H., Park, H., and Listgarten, J.
\newblock Conditioning by adaptive sampling for robust design.
\newblock \emph{arXiv preprint arXiv:1901.10060}, 2019.

\bibitem[Cao et~al.(2019)Cao, Wang, and Kitani]{cao2019learnable}
Cao, S., Wang, X., and Kitani, K.~M.
\newblock Learnable embedding space for efficient neural architecture
  compression.
\newblock \emph{arXiv preprint arXiv:1902.00383}, 2019.

\bibitem[Fialho et~al.(2010)Fialho, Schoenauer, and
  Sebag]{fialho_fitness-auc_2010}
Fialho, Ã., Schoenauer, M., and Sebag, M.
\newblock Fitness-{AUC} bandit adaptive strategy selection vs. the probability
  matching one within differential evolution: an empirical comparison on the
  bbob-2010 noiseless testbed.
\newblock In \emph{Proceedings of the 12th annual conference comp on {Genetic}
  and evolutionary computation - {GECCO} '10}, pp.\  1535, Portland, Oregon,
  USA, 2010. ACM Press.
\newblock ISBN 978-1-4503-0073-5.
\newblock \doi{10.1145/1830761.1830770}.
\newblock URL \url{http://portal.acm.org/citation.cfm?doid=1830761.1830770}.

\bibitem[G{\'o}mez-Bombarelli et~al.(2018)G{\'o}mez-Bombarelli, Wei, Duvenaud,
  Hern{\'a}ndez-Lobato, S{\'a}nchez-Lengeling, Sheberla, Aguilera-Iparraguirre,
  Hirzel, Adams, and Aspuru-Guzik]{gomez2018automatic}
G{\'o}mez-Bombarelli, R., Wei, J.~N., Duvenaud, D., Hern{\'a}ndez-Lobato,
  J.~M., S{\'a}nchez-Lengeling, B., Sheberla, D., Aguilera-Iparraguirre, J.,
  Hirzel, T.~D., Adams, R.~P., and Aspuru-Guzik, A.
\newblock Automatic chemical design using a data-driven continuous
  representation of molecules.
\newblock \emph{ACS central science}, 4\penalty0 (2), 2018.

\bibitem[Gupta \& Kundaje(2019)Gupta and Kundaje]{gupta2019targeted}
Gupta, A. and Kundaje, A.
\newblock Targeted optimization of regulatory dna sequences with neural editing
  architectures.
\newblock \emph{bioRxiv}, pp.\  714402, 2019.

\bibitem[Gupta \& Zou(2018)Gupta and Zou]{gupta2018feedback}
Gupta, A. and Zou, J.
\newblock Feedback {GAN (FBGAN)} for {DNA}: A novel feedback-loop architecture
  for optimizing protein functions.
\newblock \emph{arXiv preprint arXiv:1804.01694}, 2018.

\bibitem[Killoran et~al.(2017)Killoran, Lee, Delong, Duvenaud, and
  Frey]{killoran2017generating}
Killoran, N., Lee, L.~J., Delong, A., Duvenaud, D., and Frey, B.~J.
\newblock Generating and designing dna with deep generative models.
\newblock \emph{arXiv preprint arXiv:1712.06148}, 2017.

\bibitem[Kingma \& Welling(2014)Kingma and Welling]{vae}
Kingma, D.~P. and Welling, M.
\newblock Auto-encoding variational {B}ayes.
\newblock In \emph{2nd International Conference on Learning Representations,
  {ICLR} 2014, Banff, AB, Canada, April 14-16, 2014, Conference Track
  Proceedings}, 2014.

\bibitem[Kusner et~al.(2017)Kusner, Paige, and
  Hern{\'a}ndez-Lobato]{kusner2017grammar}
Kusner, M.~J., Paige, B., and Hern{\'a}ndez-Lobato, J.~M.
\newblock Grammar variational autoencoder.
\newblock In \emph{Proceedings of the 34th International Conference on Machine
  Learning-Volume 70}. JMLR. org, 2017.

\bibitem[Le et~al.(2018)Le, Shimko, Aditham, Keys, Longwell, Orenstein, and
  Fordyce]{le2018comprehensive}
Le, D.~D., Shimko, T.~C., Aditham, A.~K., Keys, A.~M., Longwell, S.~A.,
  Orenstein, Y., and Fordyce, P.~M.
\newblock Comprehensive, high-resolution binding energy landscapes reveal
  context dependencies of transcription factor binding.
\newblock \emph{Proceedings of the National Academy of Sciences}, 115\penalty0
  (16):\penalty0 E3702--E3711, 2018.

\bibitem[Luo et~al.(2018)Luo, Tian, Qin, Chen, and Liu]{luo2018neural}
Luo, R., Tian, F., Qin, T., Chen, E., and Liu, T.-Y.
\newblock Neural architecture optimization.
\newblock In \emph{Advances in neural information processing systems}, pp.\
  7816--7827, 2018.

\bibitem[Miyazawa \& Jernigan(1996)Miyazawa and Jernigan]{miyazawa1996residue}
Miyazawa, S. and Jernigan, R.~L.
\newblock Residue--residue potentials with a favorable contact pair term and an
  unfavorable high packing density term, for simulation and threading.
\newblock \emph{Journal of molecular biology}, 256\penalty0 (3), 1996.

\bibitem[Pedregosa et~al.(2011)Pedregosa, Varoquaux, Gramfort, Michel, Thirion,
  Grisel, Blondel, Prettenhofer, Weiss, Dubourg, Vanderplas, Passos,
  Cournapeau, Brucher, Perrot, and Duchesnay]{scikit-learn}
Pedregosa, F., Varoquaux, G., Gramfort, A., Michel, V., Thirion, B., Grisel,
  O., Blondel, M., Prettenhofer, P., Weiss, R., Dubourg, V., Vanderplas, J.,
  Passos, A., Cournapeau, D., Brucher, M., Perrot, M., and Duchesnay, E.
\newblock Scikit-learn: Machine learning in {P}ython.
\newblock \emph{Journal of Machine Learning Research}, 12, 2011.

\bibitem[Real et~al.(2019)Real, Aggarwal, Huang, and Le]{real2019regularized}
Real, E., Aggarwal, A., Huang, Y., and Le, Q.~V.
\newblock Regularized evolution for image classifier architecture search.
\newblock In \emph{Proceedings of the AAAI Conference on Artificial
  Intelligence}, volume~33, 2019.

\bibitem[Roeder et~al.(2018)Roeder, Killoran, Grathwohl, and
  Duvenaud]{roeder2018design}
Roeder, G., Killoran, N., Grathwohl, W., and Duvenaud, D.
\newblock Design motifs for probabilistic generative design.
\newblock 2018.

\end{thebibliography}
\bibliographystyle{icml2020}


\end{document}


\twocolumn[
\icmltitle{Population-Based Black-Box Optimization for Biological Sequence Design: Supplementary Material}

\icmlsetsymbol{equal}{*}

\begin{icmlauthorlist}
\icmlauthor{Christof Angermueller}{goo}
\icmlauthor{David Belanger}{goo}
\icmlauthor{Andreea Gane}{goo}
\icmlauthor{Zelda Mariet}{goo}
\icmlauthor{David Dohan}{goo}
\icmlauthor{Kevin Murphy}{goo}
\icmlauthor{Lucy Colwell}{goo,uoc}
\icmlauthor{D Sculley}{goo}
\end{icmlauthorlist}

\icmlaffiliation{goo}{Google Research}
\icmlaffiliation{uoc}{University of Cambridge}

\icmlcorrespondingauthor{Christof Angermueller}{christofa@google.com}

\icmlkeywords{Machine Learning, Black-box optimization, Protein design, Protein engineering}

\vskip 0.3in
]



\printAffiliationsAndNotice{}  

\appendix
\section{In-Silico Design problems}
\label{sec:app-problems}

\addtolength{\tabcolsep}{-4pt}
\begin{table}[h]
    \centering
    \begin{tabular}{lcccccc}
    \toprule
    \multirow{2}{*}{problem} &  Num &  Num &  Batch &  Vocab & Seq & Init \\
    & inst. & rounds & size & size & length & size \\
\midrule
TfBind8 & 12 & 10 & 100 & 4 & 8 & 0 \\
TfBind10 & 2 & 10 & 100 & 4 & 10 & 0 \\
RandomMLP & 16 & 10 & 100 & 20 & 20/40 & 0 \\
RandomRNN & 12 & 10 & 100 & 20 & 20/40 & 0 \\
PfamHMM & 24 & 10 & 500 & 20 & 50-100 & 500 \\
ProteinDistance & 24 & 6 & 500 & 20 & 50-100 & 0\\
PDBIsing & 10 & 10 & 500 & 20 & 20/50 & 0 \\
UTR & 1 & 10 & 1000 & 4 & 50 & 0\\
\bottomrule
\end{tabular}
    \caption{Information about benchmark problems, including the number of instances per problem, the number of optimization rounds, batch size, vocabulary size, sequence length, and the number of initial samples. RandomMLP and RandomRNN considers sequences of length 20 or 40. The sequence length of PfamHMM and ProteinDistance varies between 50 and 100. PfamHMM provides methods with 500 initial samples. All other problems do not provide initial samples.
    }
    \label{tab:app-problems}
\end{table}

Table \ref{tab:app-problems} summarizes the considered optimization problems. For each problem, we construct multiple instances by varying the protein target (PDBIsing, PfamHMM, TfBind), or the neural network architecture and random seed for weight initialization (RandomMLP/RandomRNN). We provide additional details per problem in the following sub-sections.

\subsection{TfBind}
\label{sec:app-problems-tfbind}
The optimization goal is to produce length-8 DNA sequences that maximize the binding affinity towards a particular transcription factor. We use the following transcription factor to create 12 optimization problems: CRX\_REF\_R1, CRX\_R90W\_R1, NR1H4\_REF\_R1, NR1H4\_C144R\_R1, HOXD13\_REF\_R1, HOXD13\_Q325R\_R1, GFI1B\_REF\_R1, FOXC1\_REF\_R1, PAX4\_REF\_R1,  PAX4\_REF\_R2, POU6F2\_REF\_R1, and SIX6\_REF\_R1.

We min-max normalize the binding affinity values for each transcription factor target to the zero-one interval.

TfBind10 differs from TfBind8 in that sequences are of length 10 and only two transcription factors (Cbf1 and Pho4) are available, which were characterized experimentally~\cite{le2018comprehensive}.

\subsection{Random Neural Network}
The goal is to optimize the scalar output of a randomly initialize neural network. Optimization proceeds over 10 rounds with a batch size 100. We construct different instances by varying the sequence length (20 or 40), the vocabulary size (4 or 20), the random seed (0 or 13), and the architecture of the network (described in the following). 

RandomMLP considers networks with a varying number of convolutional and fully connected layers. Networks have either no convolutional layer, or one layer with 128 units, a kernel width of 13, and a stride size of 1. The number of fully connected layers is either one (128 hidden units) or three (128, 256, 512 hidden units). We use a linear activation function for the output layer and a relu activation function for all other layers.

RandomRNN considers networks with one (128 hidden units), two (128, 256 hidden units), or three (128, 256, 512 hidden units) LSTM layers.

\subsection{PfamHMM}

Sequences annotated by Pfam within each family have variable length and the likelihood under the HMM can be evaluated for arbitrary-length sequences. For simplicity, however, here we consider optimization over fixed-length sequences. The length is chosen as the median length of unaligned sequence domains that belong to the Pfam-full sequence alignment for the corresponding family. 

The initial dataset is obtained by (1) selecting all sequences from Pfam-full that belong to the given family and have the chose length, (2) evaluating their likelihood under the HMM, and (3) sampling 500 sequences with a likelihood below the 50th percentile.

We selected families relatively short sequences. Future work might consider longer sequences while optimizing only a subset of positions.

\subsection{ProteinDistance}
The ProteinDistance problem tasks methods to find sequences with a high cosine similarity in the embedding space to a chosen target sequence. We used the network from \citet{bileschi2019using} to obtain embeddings of proposed sequences. This network was trained to classify the Pfam family of a protein sequence, and its embeddings have been shown to capture broad protein features that are useful for fewshot learning. We construct optimization problems by choosing different Pfam HMM seed sequences as the target sequence, and used the same Pfam families that we used the PfamHMM problem.

\subsection{PDBIsingModel}
We employ $\fx = \sum_i \phi_i(x_i) + \beta \sum_{ij} C_{ij} \phi(x_i,  x_j)$, where $x_i$ refers to the character in the $i$-th position of sequence $x$. $C_{ij}$ is a binary \textit{contact map} dictating which positions are in contact and $\phi(x_i,  x_j)$ is a position-independent \textit{coupling block}. We construct optimization problems by choosing 10 different proteins from the Protein Data Bank~\citep{berman2003protein}. $C_{ij} = 1$ if the $C\alpha$ atoms of the amino acid residues at positions $i$ and $j$ are separated by less than 8 Angstroms in the protein's 3D structure. The coupling block is based on global co-occurence probabilities of contacting residues \citep{miyazawa1996residue}. The local terms $\phi_i(x_i)$ are set using the sequence of amino acids for the true PDB protein. The score for setting $x_i$ to a specified value is given by the log Blosum substitution probability between that value and the corresponding value in the PDB sequence. Finally, $\beta$ is chosen heuristically such that the local terms do not dominate the objective too much.

\section{Optimization Methods}
\label{sec:app-methods}

\subsection{Model-Based Optimization}
\label{sec:app-mbo}
We follow~\cite{Angermueller2020RLDesign} for building the regressor model. We optimize the hyper-parameters of diverse regressor models by randomized search, and evaluate model performance by explained variance score estimated by five-fold cross validation. We select all models with a score $\ge0.4$ and build an ensemble by averaging their predictions. We consider the following scikit-learn~\cite{scikit-learn} regressor classes and hyper-parameters:
\begin{itemize}[topsep=0pt,noitemsep]
    \item BayesianRidge: alpha\_1, alpha\_2, lambda\_1, lamdba\_2
    \item RandomForestRegressor: max\_depth, max\_features, n\_estimators
    \item LassoRegressor: alpha
    \item GaussianProcessRegressor: kernel (RBF, RationalQuadratic, Matern) and kernel parameters
\end{itemize}

We use the posterior mean acquisition function, which performed as good or better as the upper confidence bound, expected improvement, or probability of improvement.
We optimize the acquisition function using evolutionary search for 500 rounds with a batch size of 25. We construct the next batch by selecting the top $n$ unique and novel sequences with the highest acquisition function value.

\subsection{Latent-Space Model-Based Optimization}

An alternative approach to discrete optimization is to train an encoder-decoder model that enables mapping discrete sequences to and from continuous vectors and moving the optimization process into the continuous space \cite{gomez2018automatic, kusner2017grammar, roeder2018design, killoran2017generating, cao2019learnable, luo2018neural, gupta2019targeted}. The approach has been introduced for problems with a large amount of (unlabeled) initial data in a single-round optimization set up, which we extend to multi-round optimization. In each round, we use all the observed sequence-reward pairs to jointly train a variational auto-encoder (VAE) \cite{vae} and a neural network regressor from latent embeddings to the corresponding rewards. Jointly training the regressor and VAE encourages organizing latent representations by reward scores \cite{gomez2018automatic}. Once the encoder-decoder model has been fixed, similar to the previous work, we train a new regressor from scratch on the embedding-reward pairs, and use it to score new sequences during continuous optimization.

We train the VAE by maximizing the variational lower-bound \cite{vae}. We anneal the KL divergence and the neural network regressor loss over time. We standard scale regressor target labels, and use the mean-squared error as traning objective. The encoder and decoder architectures match those in the DbAs-VAE; the regressor is a fully-connected network with a single hidden layer. We consider the following hyper-parameters (actual values in the parentheses): latent space size (50), learning rate (0.006), weight of the regressor loss term (2 * sequence length), sigmoid annealing slope (1.), batch size (20), number of epochs (60), hidden sizes for the encoder (128), decoder (128), and regressor (50). 

For training a new regressor on the embedding-reward pairs, we use the ensemble-based approach described Section~\ref{sec:app-mbo}.

We perform optimization in the latent space using the cross-entropy method with a diagonal multivariate Gaussian generative model. We perform weighted maximum likelihood on high scoring samples, re-weighted by the likelihood ratio of a given prior and the current generative model (CbAs) \cite{brookes2019conditioning}. We use as prior a diagonal multivariate Gaussian generative model fit on the training data of the VAE and regressor. The re-weighting approach encourages the optimization trajectory to remain close to the prior, which is desirable as the latent regressor performance is expected to degrade as we move away from training samples.

We run the cross-entropy method for a fixed number of iterations. The proposed batch is obtained by decoding the best-scoring unique samples (according to the regressor score) across multiple runs of the cross-entropy method with different random seeds. We tune the following hyper-parameters (actual values in the parentheses): the number of cross-entropy method instances (25), the number iterations per instance (20), the number of samples drawn per iteration (100), and the number high-scoring samples to extract (10). We initialize the cross-entropy method with the embeddings of the highest scoring 25 sequences observed during the optimization. 
Finally, we note that, as optimization is being performed in the continuous space, one could use a differentiable regressor and gradient-based optimization.

\subsection{Evolution}
Evolution generates a new child sequence by selecting two parents from the population of $(\x, \fx)$ pairs, recombining them, and mutating them. At each optimization rounds, at repeats these steps iteratively to generate a batch of child sequences. Following \cite{real2019regularized}, samples are no longer considered during parent selection after a fixed number optimization rounds to prevent elite samples from dominating the population ("death by old age").  Two parents are chosen for each child via tournament selection, which involves taking the best of T samples from the alive population.  The chosen parent sequences A and B are recombined by copying the sequences from left to right beginning with a pointer on parent A. At each position, the pointer has some probability (we used 0.1) of switching to reading the other parent. After crossover, we mutate the child by changing each position to a different value with a fixed mutation probability (we used 0.01).

\subsection{DbAs-VAE}
We follow~\cite{brookes2018design} and use a fully connected MLP encoder and decoder with one hidden layer and 64 units, 32 latent vectors, which is trained with a learning rate of 0.01 for 60 epochs and a batch size of 20. We use a quantile-cutoff of 0.85.

\subsection{DbAs-RNN}
Same as DbAs-VAE, except that the generative model is a LSTM with one hidden layer and 128 hidden units.

\subsection{FBGAN}
Following~\cite{gupta2018feedback}, we use a Wasserstein GAN as generative model inside cross-entropy optimization. Unlike using a fixed threshold for selecting sequences, we use a quantile cutoff since it performed better in our experiments on diverse problems. We tune the quantile cutoff, learning rate, batch size, discriminator and generator training epochs, the gradient penalty weight, the Gumble softmax temperature, and the number of latent variables of the generator.

\subsection{\pbo}
\label{sec:app-pbo}
\begin{figure*}[tb]
\includegraphics[width=\textwidth]{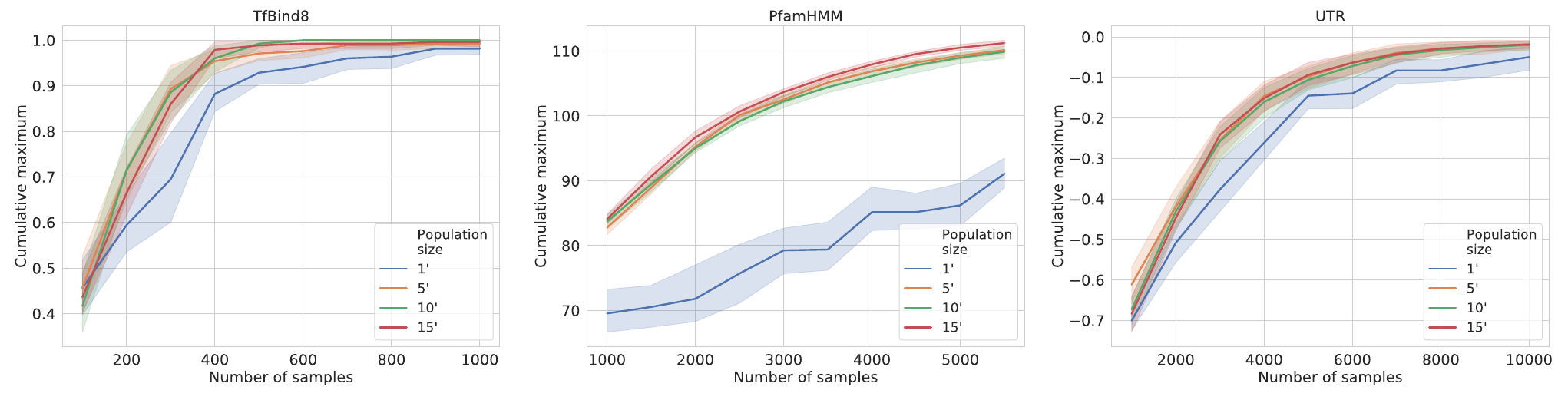}
\caption{Cumulative maximum of \pbo for different population sizes $N$ on three optimization problems. While a population size of 15 is best on average, similar performances can be achieved with smaller populations. Using only one algorithm ($N=1$) results in a clear performance decrease. For this analysis, the initial population was sampled randomly from a pool of algorithms as described in \ssecref{sec:exp-pbo}{6.2}, subject to sampling at least one instance per algorithm class.}
\label{fig:psize}
\end{figure*}

We use the population size of 15, and note that similar performance can be achieved with a smaller population size greater than 5 (see \figref{fig:psize}). We use MBO, DbAs-VAE, DbAs-RNN, Evolution, and SMW as constituent algorithm classes (\ssecref{sec:exp-methods}{6.1}). We sample hyper-parameters of these algorithm classes from following distributions:

\paragraph{SMW}
This method is hyper-parameter free.

\paragraph{Evolution}
\begin{itemize}[topsep=0pt,noitemsep]
    \item crossover\_probability: uniform(interval(0.1, 0.3))
    \item mutation\_probability: uniform(interval(0.05, 0.2))
\end{itemize}

\paragraph{DbAs-VAE}
\vspace{-.8em}
\begin{itemize}[topsep=0pt,noitemsep]
    \item quantile: uniform(0.825, 0.975)
    \item learning\_rate: loguniform(interval(0.008, 0.012))
    \item num\_vae\_units: uniform(categorical(64, 128))
\end{itemize}

\paragraph{DbAs-RNN}
\vspace{-.8em}
\begin{itemize}[topsep=0pt,noitemsep]
    \item quantile: uniform(0.9, 0.975)
    \item learning\_rate: loguniform(interval(0.0005, 0.012))
    \item num\_lstm\_units: uniform(categorical(64, 128))
\end{itemize}

\paragraph{MBO}
\vspace{-.8em}
\begin{itemize}[topsep=0pt,noitemsep]
    \item acquisition\_function(uniform(categorical([‘PosteriorMean’, ‘UCB’])))
    \item ucb\_scale\_factor: uniform(interval(0.5, 1.2))
    \item \begin{flushleft} regressor: uniform(categorical(['Ensemble', 'BayesianRidge']))\end{flushleft}
\end{itemize}

Here, 'Ensemble' refers to the ensemble model described in \ref{sec:app-mbo} and 'BayesianRidge' to the BayesianRidge regressor of the scikit-learn package.

\section{Diversity metrics}
\label{sec:app-diversity}

Finding not only one but multiple diverse high reward sequences is desirable to increase the chance that some of the found sequences satisfy downstream screening criteria that are not captured by the primary optimization objective, e.g. stability or viscosity. For quantifying diversity, we devised the following metrics:

\paragraph{Mean pairwise Hamming distance}
We compute the mean pairwise Hamming distance of sequence in a batch. We also considered the Edit distance but found it to be too slow to compute for problems with large batch sizes.

\paragraph{Mean entropy}
For a given batch of sequences, we compute the entropy over characters at each position, and then average over positions. Since we found this metric to be highly correlated with the mean pairwise Hamming distance ($R^2 = 0.99$), we only show results for the mean pairwise Hamming distance.

\paragraph{Reward vs. Hamming distance}
We plot the mean reward of sequences in a batch. vs. the mean pairwise Hamming distance, which shows the trade-off of these two metrics.

\paragraph{Number of high-reward clusters}
For problems with known maximum we first select all sequences proposed so far with a reward above a threshold, e.g $0.8 \cdot \max f(x)$. We then cluster the selected sequences using hierarchical complete linkage clustering, and count the number of resulting clusters with a minimum length-normalized Hamming distance if 0.5.  This metric can be only maximized by finding both well separated and high reward sequences.

\paragraph{Fraction of optima found}
This metric can be only computed for enumerative optimization problems, in our case TfBind8. In a pre-computation step, we first select all possible sequences with a reward $\ge 0.9$ (the maximum reward is $1.0$, Section~\ref{sec:app-problems-tfbind}). We then cluster selected sequences into distinct clusters with a minimum edit distance of 3 using hierarchical complete linkage clustering. We define the sequence with the highest reward of each cluster as one optimum. We count an optimum as found if the exact same sequence is proposed. We treat the forward and reverse sequence as two distinct optima. This metric differs from the \textit{Number of high-reward clusters} metric in two points: 1) the number of clusters is normalized by the total number of clusters, and 2) the edit distance is used for clustering sequences to assign sequences that contain similar motifs at different positions to the same cluster.

\begin{figure*}[tb]
\includegraphics[width=\textwidth]{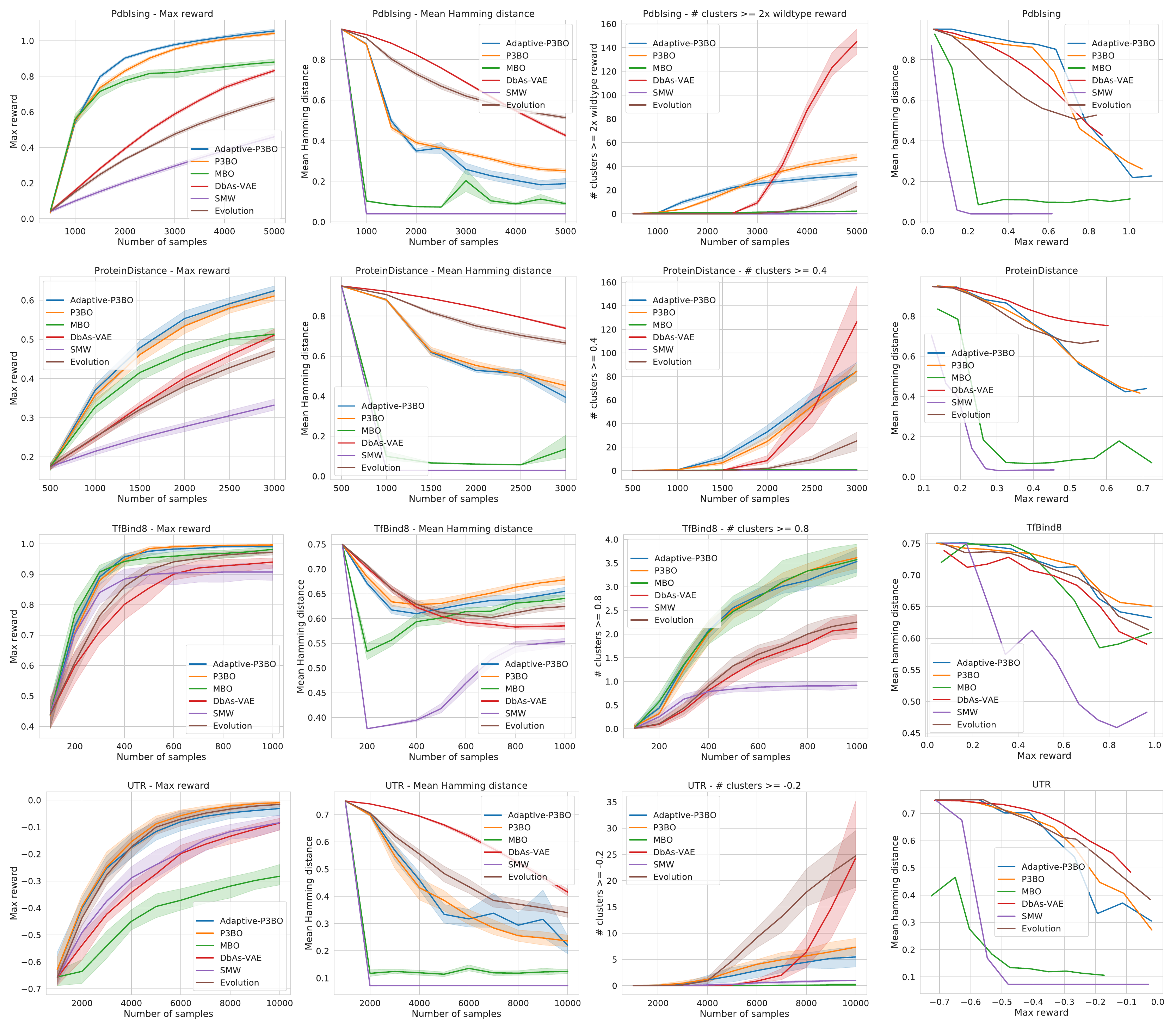}
\caption{Visualization of alternative diversity (Section~\ref{sec:app-diversity}) for the PdbIsing, ProteinDistance, TfBind8, and UTR problem. We find that DbAs-VAE and Evolution generate more diverse sequences than \pbo, albeit with a lower reward. \pbo finds significantly more diverse and higher reward sequences than both MBO and SMW, and provides overall a good trade-off between finding high reward and diverse sequences. }
\label{fig:diversity}
\end{figure*}

\section{Additional results}

\begin{figure*}[tb]
\includegraphics[width=\textwidth]{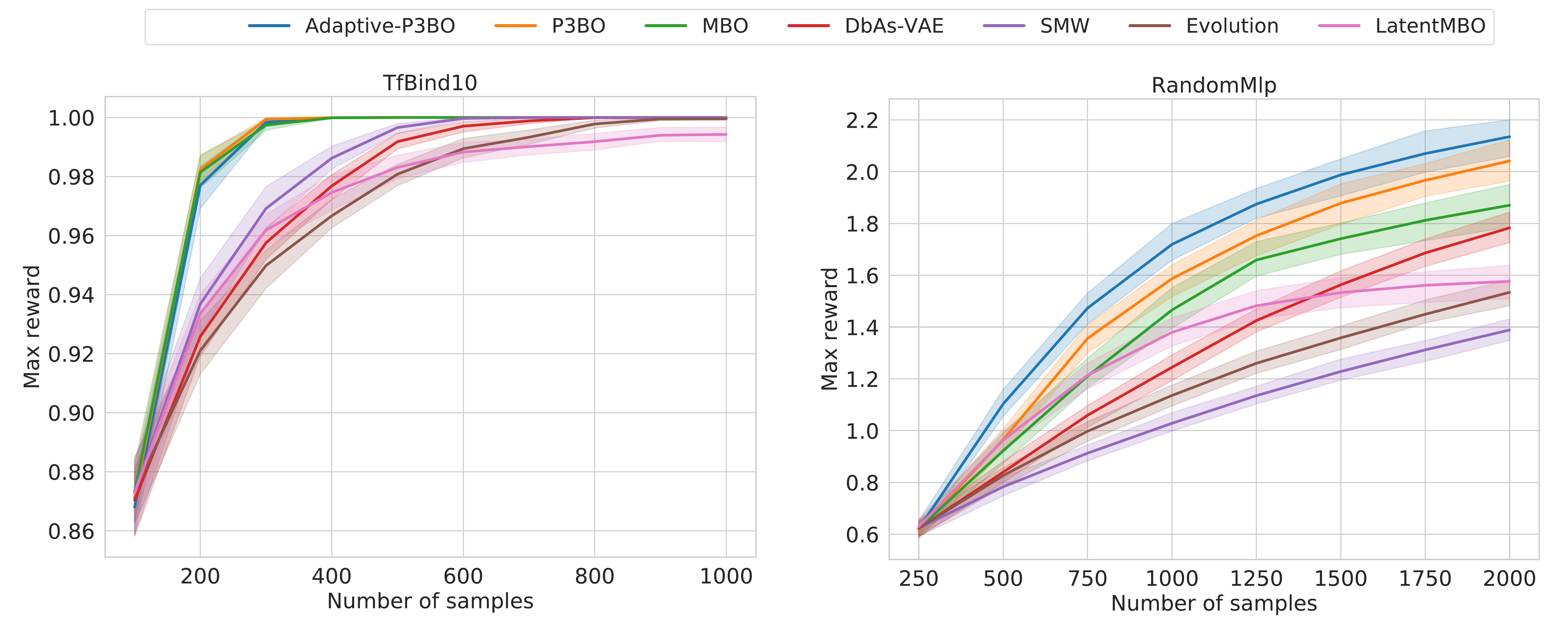}
\caption{Performance curves for the RandomMLP and TfBind10 problem, which were not shown in the main text due to space limitations. Lines show the average over all targets available for each problem, while the shaded areas indicate the $95\%$ boostrap confidence-intervals.}
\label{fig:perf_other_problems}
\end{figure*}

\begin{figure*}[tb]
\includegraphics[width=\textwidth]{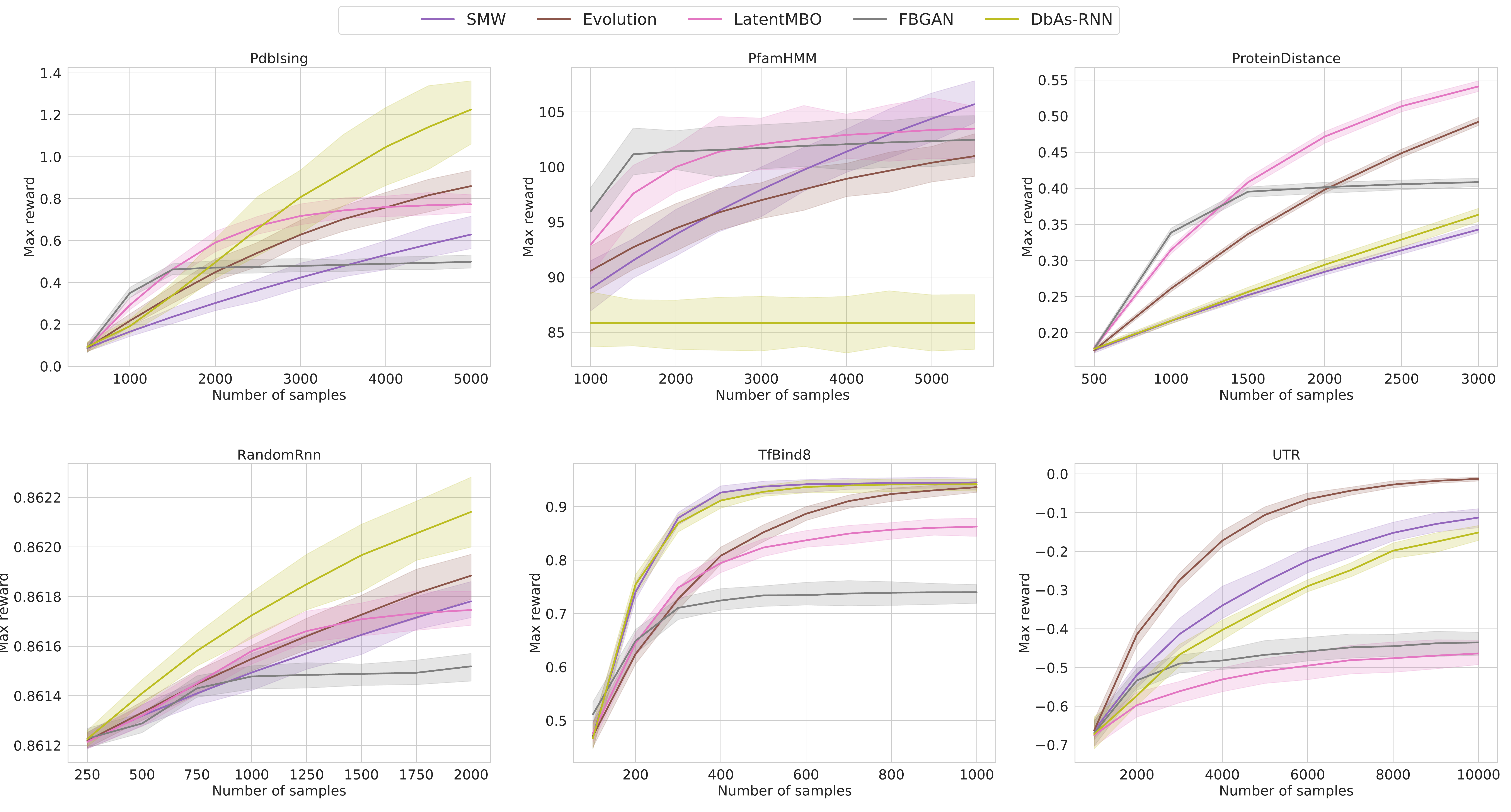}
\caption{Performances curves for additional baselines (FBGAN and DbAs-RNN), which were not shown in the main text due to space limitations.  Lines show the average over all instances per optimization problem, while the shaded areas indicate the $95\%$ bootstrap confidence-intervals.}
\label{fig:perf_baselines}
\end{figure*}

\begin{figure*}[t]
\centering
\includegraphics[width=1.0\textwidth]{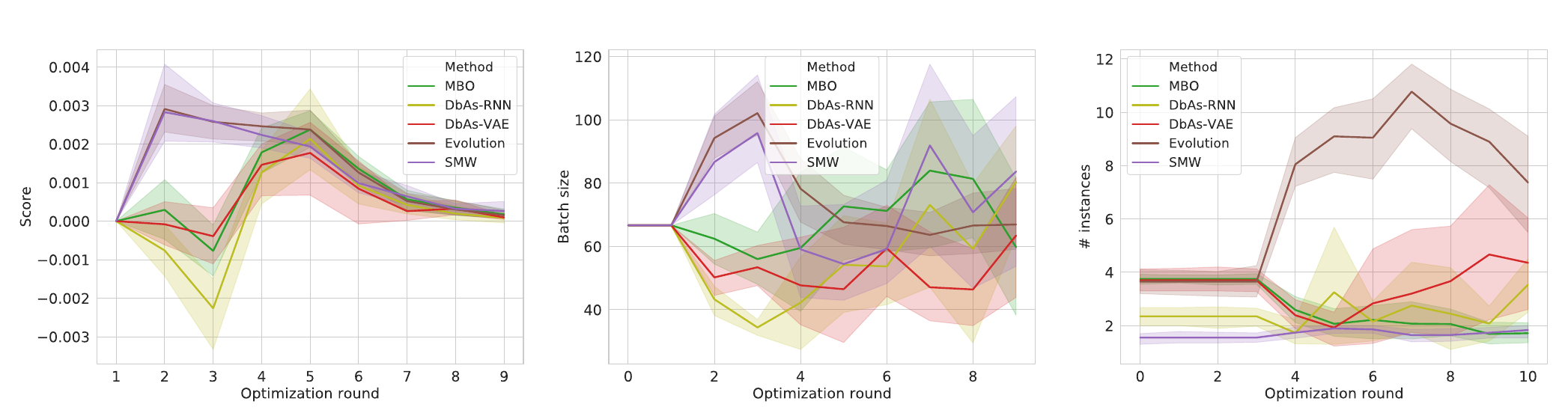}
\caption{Insights into Adaptive-\pbo applied to the UTR problem. Shown are the credit score (left), the number of sequences sampled (middle), and the number of instances (right) per algorithm class over time. Since \textit{Evolution} has the highest credit score (relative improvement) for early rounds, more sequences are sampled from \textit{Evolution} (middle), and Adaptive-\pbo increases the number of \textit{Evolution} instances from 4 to 11 (the total population size is 15). The adaptation starts after three warm-up rounds used to reliably estimate the credit score of algorithms.}
\label{fig:pbo_members_utr}
\end{figure*}

\begin{figure*}[t]
\centering
\subfigure[Performance of \pbo when removing individual algorithms.]{\includegraphics[width=\textwidth]{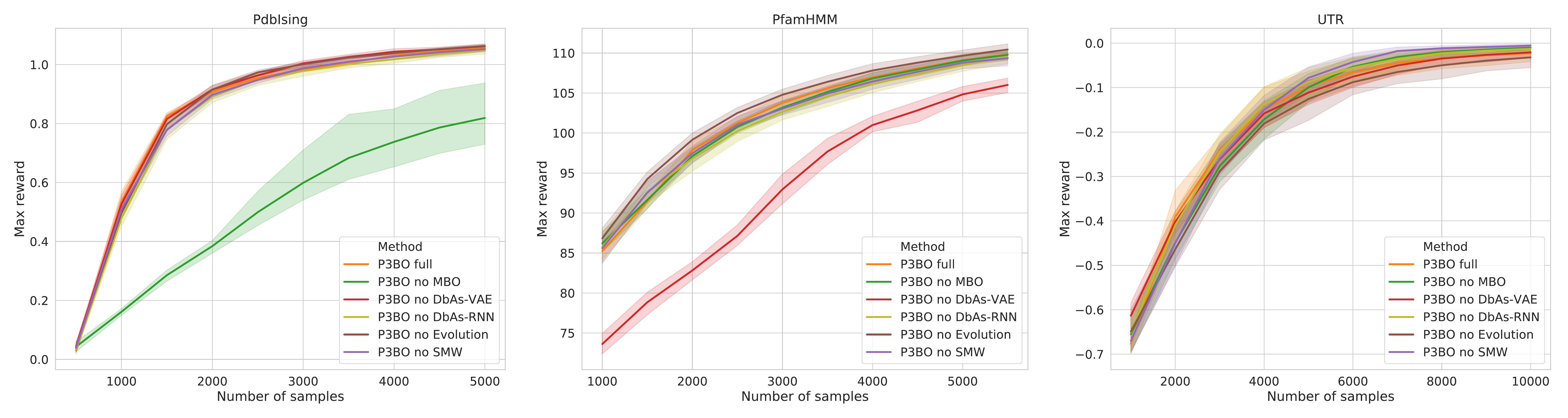}\label{fig:abl_01}}
\subfigure[Performances of methods when used stand-alone without data sharing.]{\includegraphics[width=\textwidth]{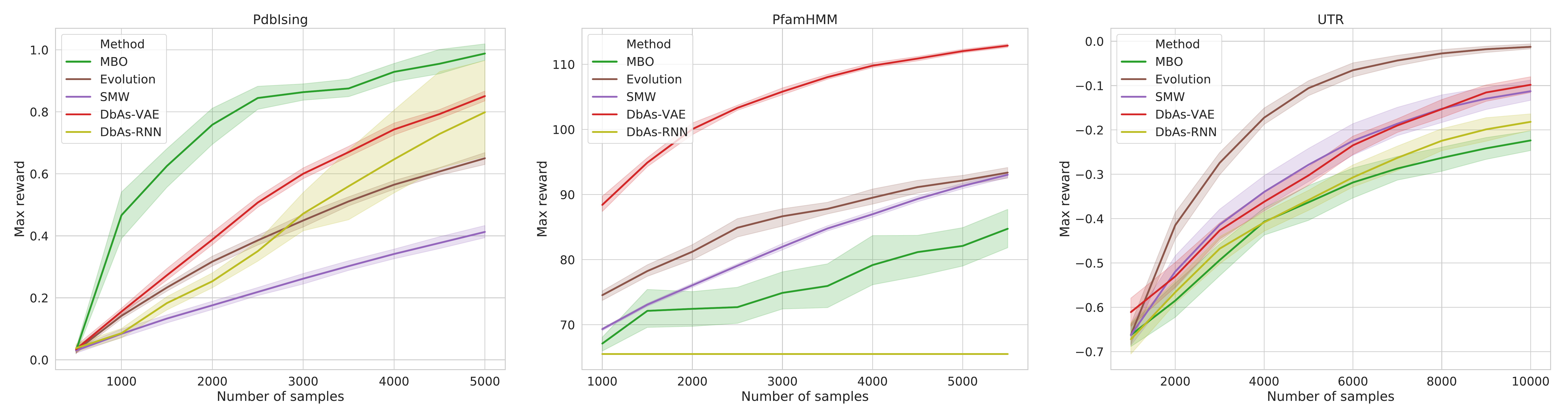}\label{fig:abl_02}}
\subfigure[Performances of methods when used inside \pbo with data sharing.]{\includegraphics[width=\textwidth]{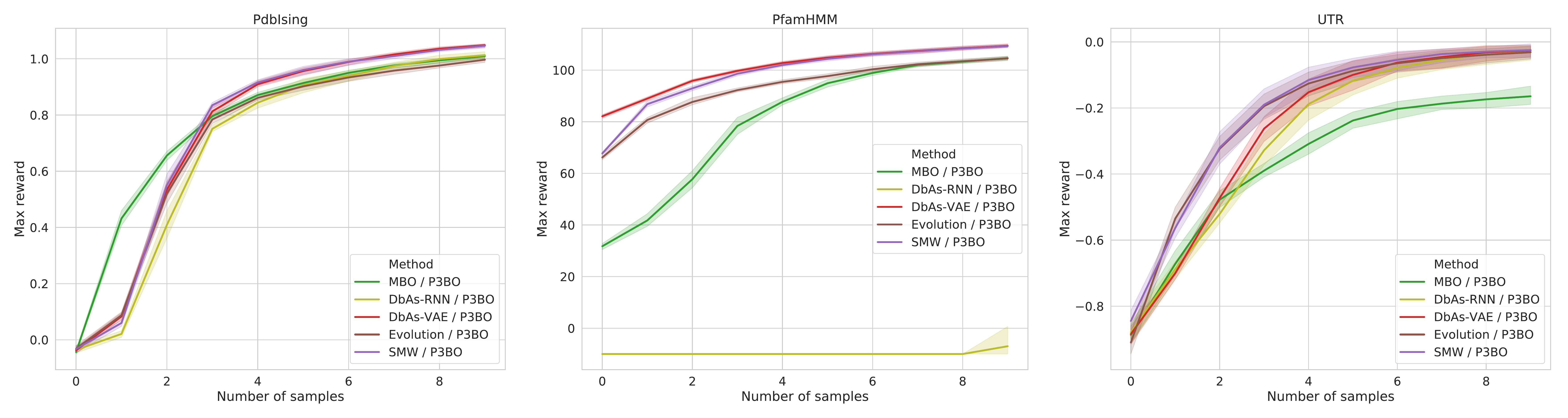}\label{fig:abl_03}}
\caption{
Performance of \pbo on the PdbIsing, PfamHMM, and UTR problem when removing individual algorithms from its population. The top row compares \pbo with all algorithms (\pbo full) to variants with one algorithm removed. The middle row shows the performance of algorithms when used stand-alone without sharing data (samples $(x, f(x))$)a, and the bottom row when used inside \pbo with data sharing. Removing MBO from the population of \pbo results in a performance drop on PdbIsing (top row) due to the good performance of MBO on that problem (middle row). In contrast, DbAs-VAE is the best performing algorithm on PfamHMM, which results in a performance drop when removing it. Sharing samples acquired by one algorithm with all other algorithms in the population (bottom row) results in a higher performance of individual algorithms than without sharing samples (middle row). For example, SMW benefits from the high-reward sequences found by MBO on PdbIsing in early rounds (middle row, left plot) and thereby manages to find sequences with a higher reward than MBO in following rounds (bottow row, left plot). 
}
\label{fig:abl}
\end{figure*}

\begin{figure*}[t]
\centering
\subfigure[\tsne of sequences without data sharing.]{\includegraphics[width=\textwidth]{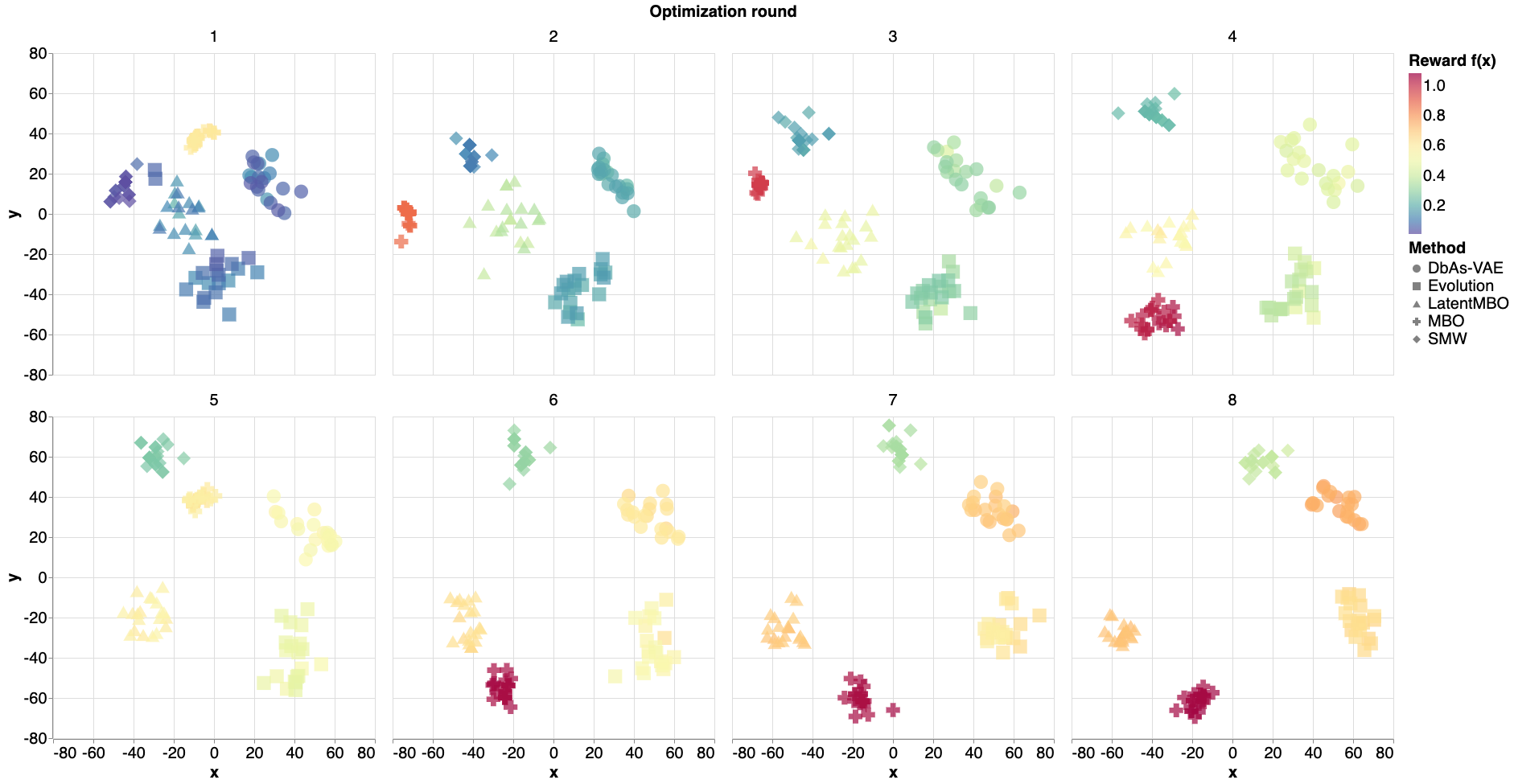}\label{fig:tnse_data_sharing_off}}
\subfigure[\tsne of sequences with data sharing.]{\includegraphics[width=\textwidth]{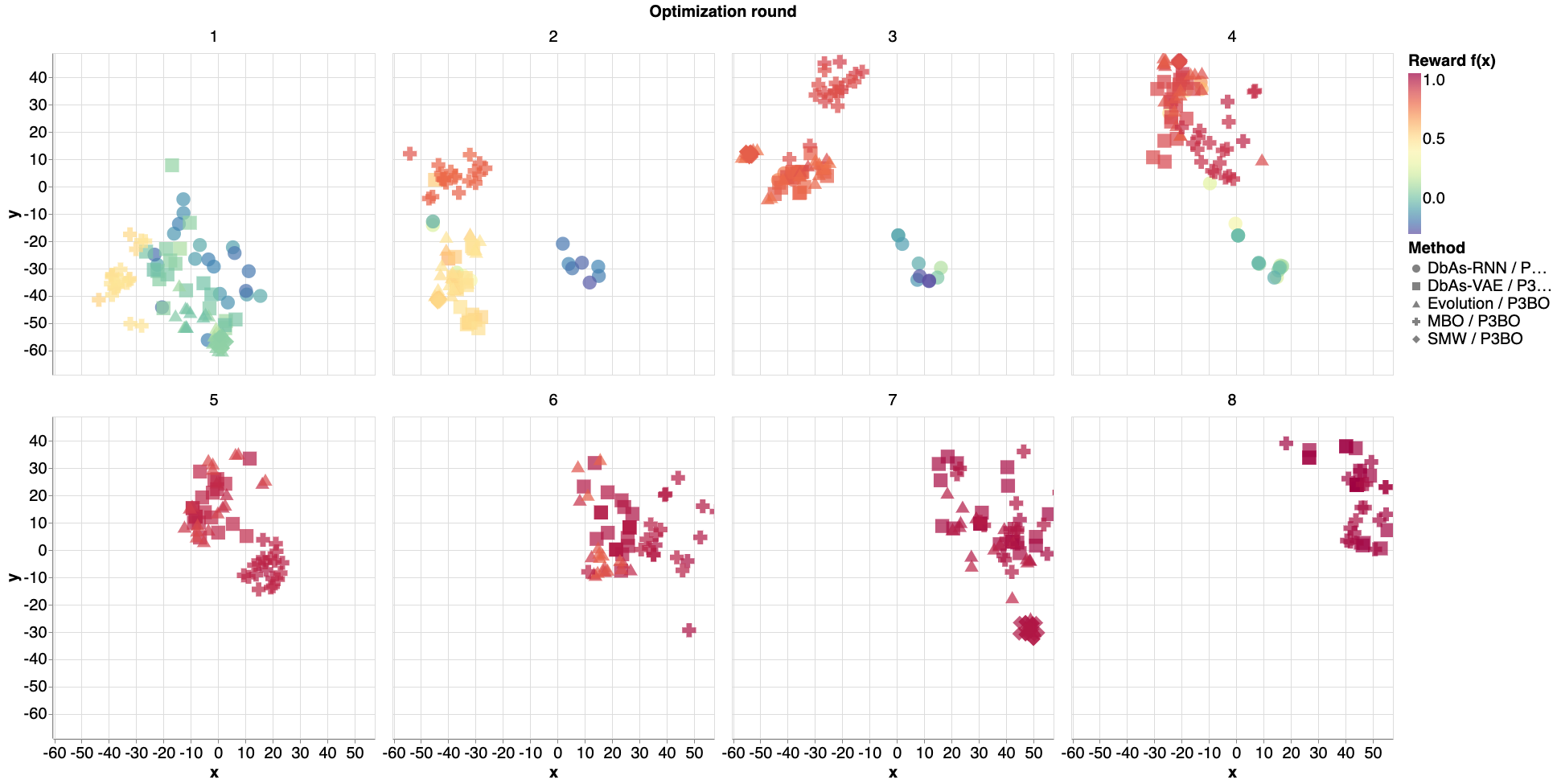}\label{fig:tsne_data_sharing_on}}
\caption{$t$-SNE of sequences proposed by different algorithms with and without sharing samples $(x, f(x))$ between algorithms. 
The shape of each point (sequence) corresponds to the algorithm that proposed the sequence $x$ and the color to the reward $f(x)$. Without sharing samples, methods propose distinct, well separated, sequences, and only MBO finds high reward sequences quickly. By sharing samples, methods benefit from the high reward sequences discovered by MBO in early rounds and propose similar sequences in subsequent rounds. Results are shown for PdbIsing problem with PDB structure 1KDQ.}
\label{fig:tsne_data_sharing}
\end{figure*}

\begin{figure*}[tb]
\includegraphics[width=\textwidth]{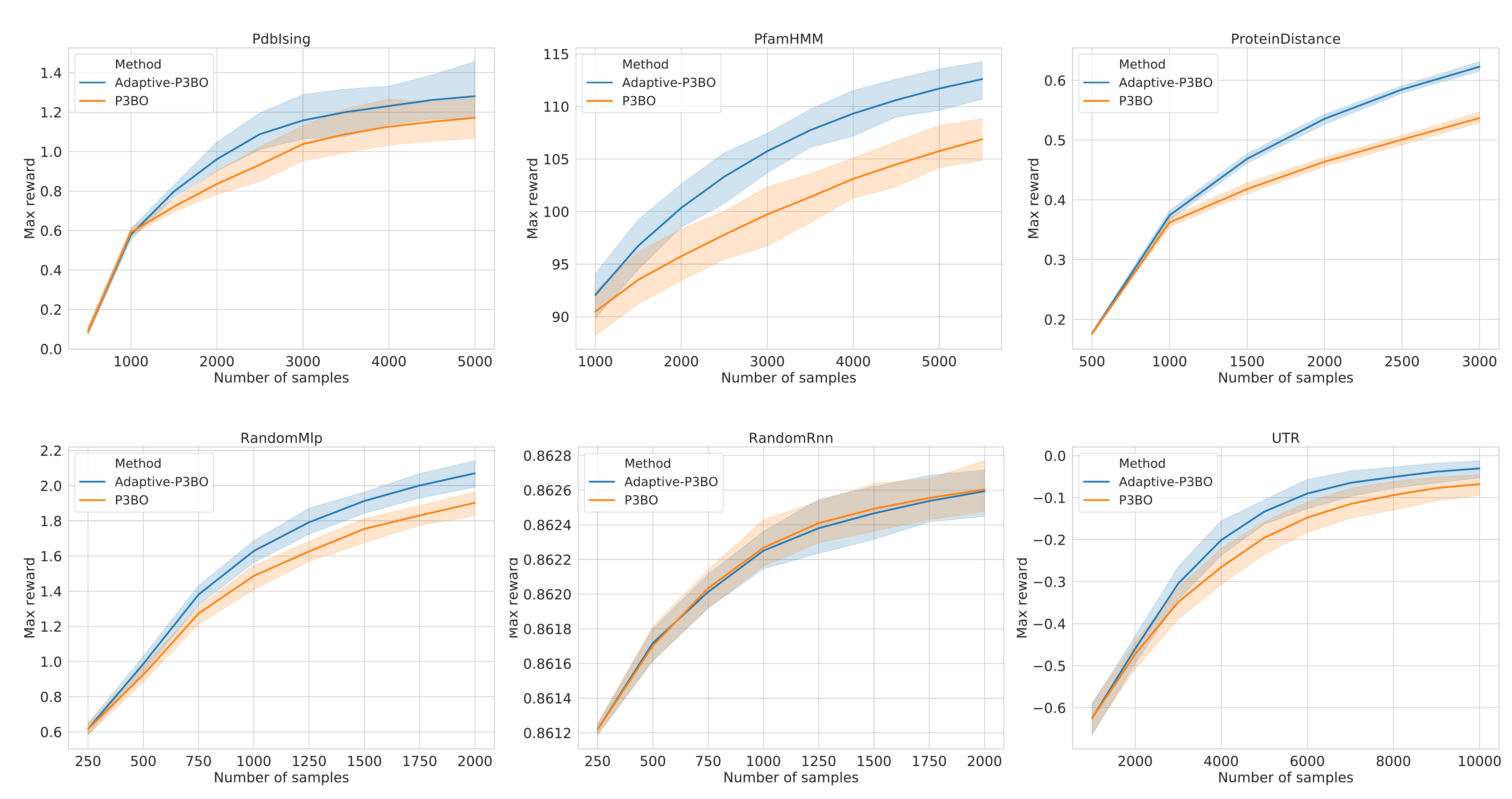}
\caption{Performance comparison of P3BO and Adaptive-\pbo when starting with a poorly initialized population of algorithms. By adapting the population online, Adaptive-P3BO can recover from the initial population, resulting in a clear performance improvement.}
\label{fig:adaptive_pbo_bad_init}
\end{figure*}

\begin{figure*}[t]
\centering
\includegraphics[width=1.0\textwidth]{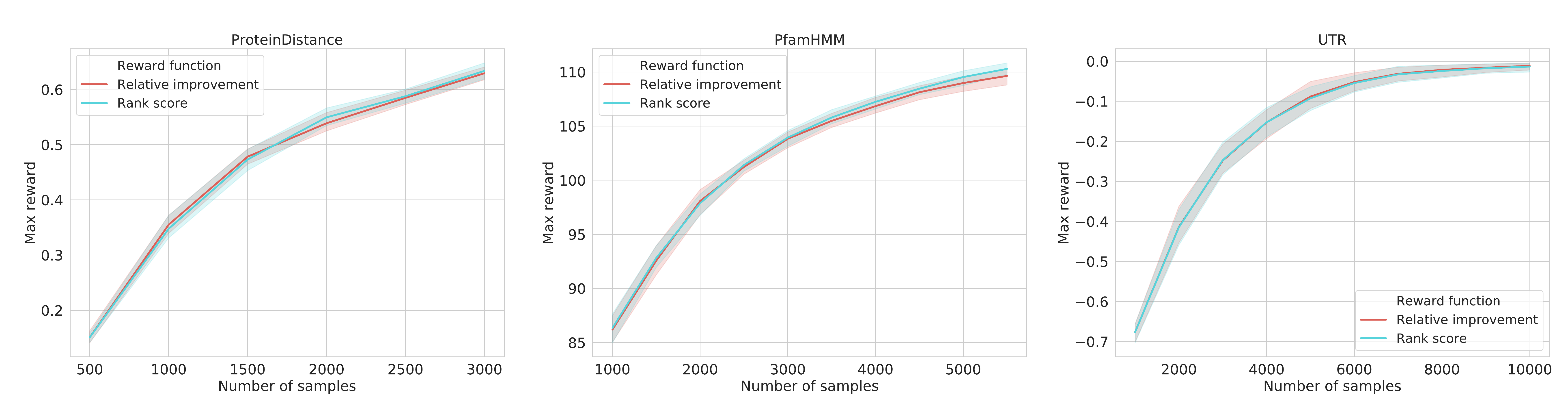}
\caption{Comparison of the relative improvement reward function described in \ssecref{sec:credit-scores}{3.2} with the rank-based reward function as proposed by~\citet{fialho_fitness-auc_2010}. Both approaches perform similarly across problems.}
\label{fig:score_random}
\end{figure*}

\begin{figure*}[tb]
\includegraphics[width=\textwidth]{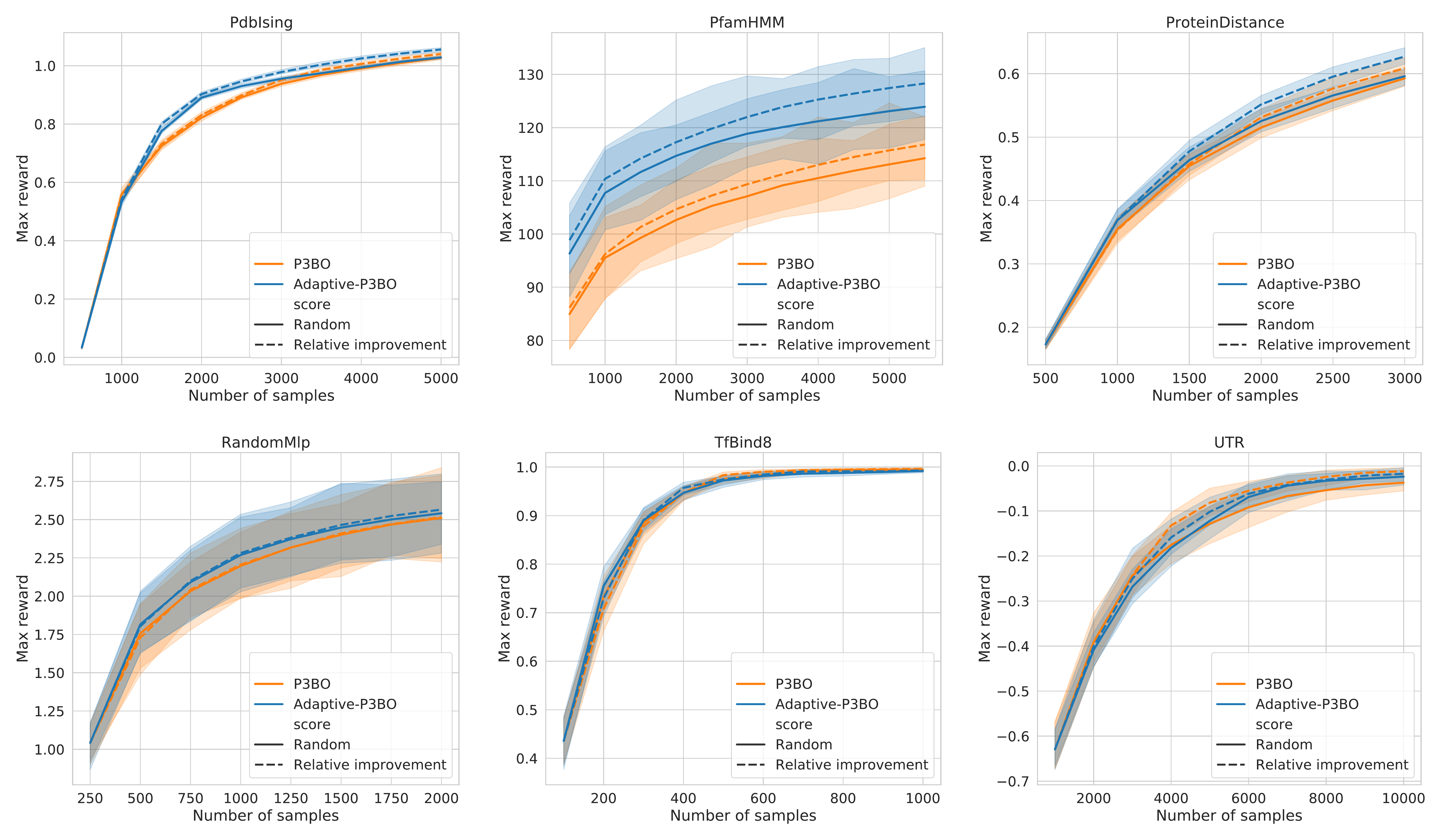}
\caption{Performance of \pbo and \apbo when using the relative improvement scoring function described in \ssecref{sec:credit-scores}{3.2} (dashed line) vs. scoring methods randomly (solid line).}
\label{fig:score_rank}
\end{figure*}

\begin{figure*}[t]
\centering
\includegraphics[width=1.0\textwidth]{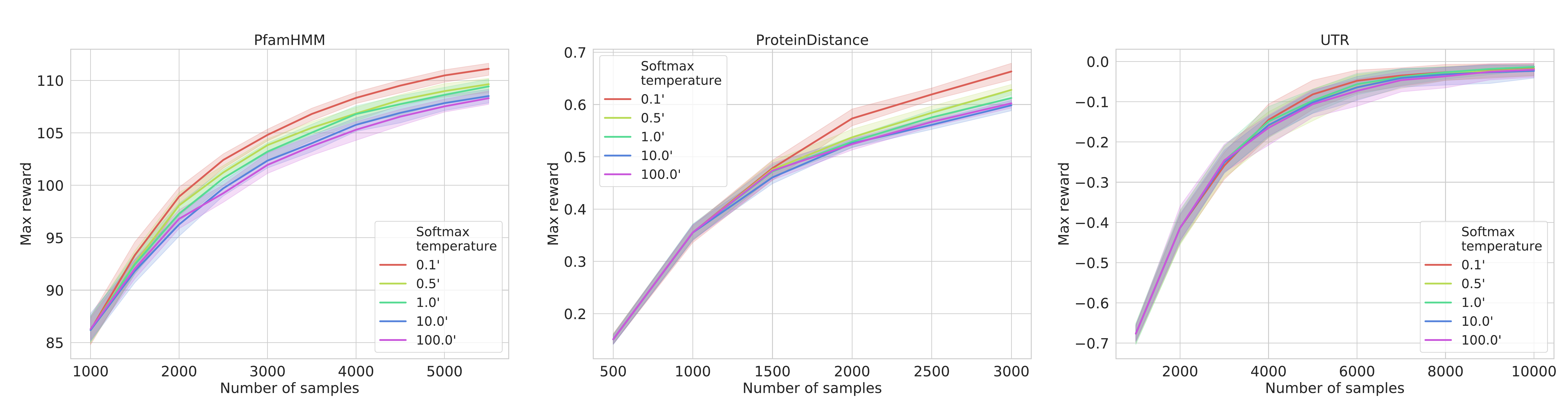}
\caption{Sensitivity of \pbo to the softmax temperature $\tau$ for computing selection probabilities (\ssecref{sec:credit-scores}{3.2}) on the PfamHMM, ProteinDistance, and UTR problem. Shows that scaling the number of sequences sampled from algorithms in the population proportional to their credit score ($\tau<10$) is better than sampling sequences uniformly ($\tau \ge 10$).}
\label{fig:score_temp}
\end{figure*}

\begin{figure*}[tb]
\includegraphics[width=\textwidth]{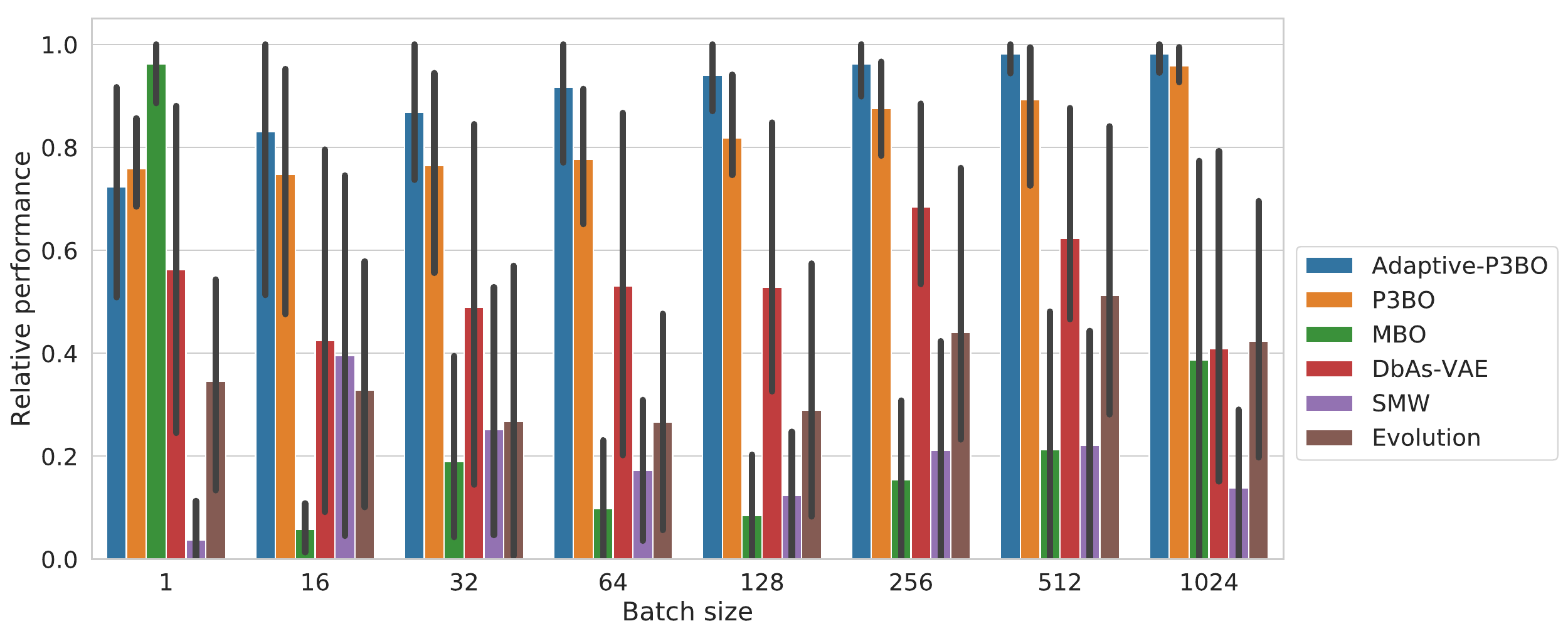}
\caption{Relative performance of methods depending on the number of samples per optimization round (batch size). The y-axis corresponds to the area under the cumulative max reward curve, min-max normalized across methods. Error bars show the variation across five optimization problems (one instance of the ProteinDistance, PfamHMM, UTR, and RandomMLP problem). Shows that \pbo is applicable to optimization settings with various batch sizes, including non-batched optimization. 
}
\label{fig:perf_vs_batch_size}
\end{figure*}

\clearpage
\bibliography{sources}
\bibliographystyle{icml2020}